\documentclass[sigconf, authorversion,nonacm]{acmart}

\usepackage{xspace}
\usepackage{multirow}
\usepackage{balance}
\usepackage{enumitem}
\usepackage{hhline}
\usepackage[noend,ruled]{algorithm2e}

\newcommand{\etal}{\textit{et al.\xspace}}

\newcommand{\eg}{\textit{e.g.}, }

\AtBeginDocument{%
  \providecommand\BibTeX{{%
    \normalfont B\kern-0.5em{\scshape i\kern-0.25em b}\kern-0.8em\TeX}}}

\usepackage{etoolbox}
\makeatletter
\patchcmd{\maketitle}{\@copyrightpermission}{}{}{}

\makeatother

\begin{document}

\title{Beyond Geo-localization: Fine-grained Orientation of Street-view Images by Cross-view Matching with Satellite Imagery}

\author{Wenmiao Hu}
\affiliation{%
  \institution{Grab-NUS AI Lab, NUS}
  \country{}
}
\email{hu.wenmiao@u.nus.edu}

\author{Yichen Zhang}
\affiliation{%
  \institution{Grab-NUS AI Lab, NUS}
  \country{}}
\email{zhang.yichen@u.nus.edu}

\author{Yuxuan Liang}
\affiliation{%
  \institution{National University of Singapore}
  \country{}
  }
\email{yuxliang@outlook.com}

\author{Yifang Yin}
\affiliation{%
  \institution{Institute for Infocomm Research, A*STAR}
  \country{}}
\email{yin_yifang@i2r.a-star.edu.sg}

\author{Andrei Georgescu}
\affiliation{%
  \institution{Grab Chronos S.R.L.}
  \country{}}
\email{andrei.georgescu@grabtaxi.com}

\author{An Tran}
\affiliation{%
  \institution{Grabtaxi Holdings Pte. Ltd.}
  \country{}}
\email{an.tran@grabtaxi.com}

\author{Hannes Kruppa}
\affiliation{%
  \institution{Grabtaxi Holdings Pte. Ltd.}
  \country{}}
\email{hannes.kruppa@grabtaxi.com}

\author{See-Kiong Ng}
\affiliation{%
  \institution{Grab-NUS AI Lab, NUS}
  \country{}}
\email{seekiong@nus.edu.sg}

\author{Roger Zimmermann}
\affiliation{%
  \institution{Grab-NUS AI Lab, NUS}
  \country{}}
\email{rogerz@comp.nus.edu.sg}

\renewcommand{\shortauthors}{Wenmiao Hu et al.}
\newcommand{\multiref}[2]{\autoref{#1}-\ref{#2}} 

\begin{abstract}
Street-view imagery provides us with novel experiences to explore different places remotely. Carefully calibrated street-view images (\eg Google Street View) can be used for different downstream tasks, \eg navigation, map features extraction. As personal high-quality cameras have become much more affordable and portable, an enormous amount of crowdsourced street-view images are uploaded to the internet, but commonly with missing or noisy sensor information. 
To prepare this hidden treasure for ``ready-to-use'' status, determining missing location information and camera orientation angles are two equally important tasks. Recent methods have achieved high performance on geo-localization of street-view images by cross-view matching with a pool of geo-referenced satellite imagery. However, most of the existing works focus more on geo-localization than estimating the image orientation. In this work, we re-state the importance of finding fine-grained orientation for street-view images, formally define the problem and provide a set of evaluation metrics to assess the quality of the orientation estimation. We propose two methods to improve the granularity of the orientation estimation, achieving 82.4\% and 72.3\% accuracy for images with estimated angle errors below 2$^\circ$ for CVUSA and CVACT datasets, corresponding to 34.9\% and 28.2\% absolute improvement compared to previous works. Integrating fine-grained orientation estimation in training also improves the performance on geo-localization, giving top 1 recall 95.5\%/85.5\% and 86.8\%/80.4\% for orientation known/unknown tests on the two datasets.

\end{abstract}

%
%



\maketitle
\begin{figure}
  
  \includegraphics[width=0.47 \textwidth, trim = 3.7cm 8.5cm 22.9cm 3cm, clip]{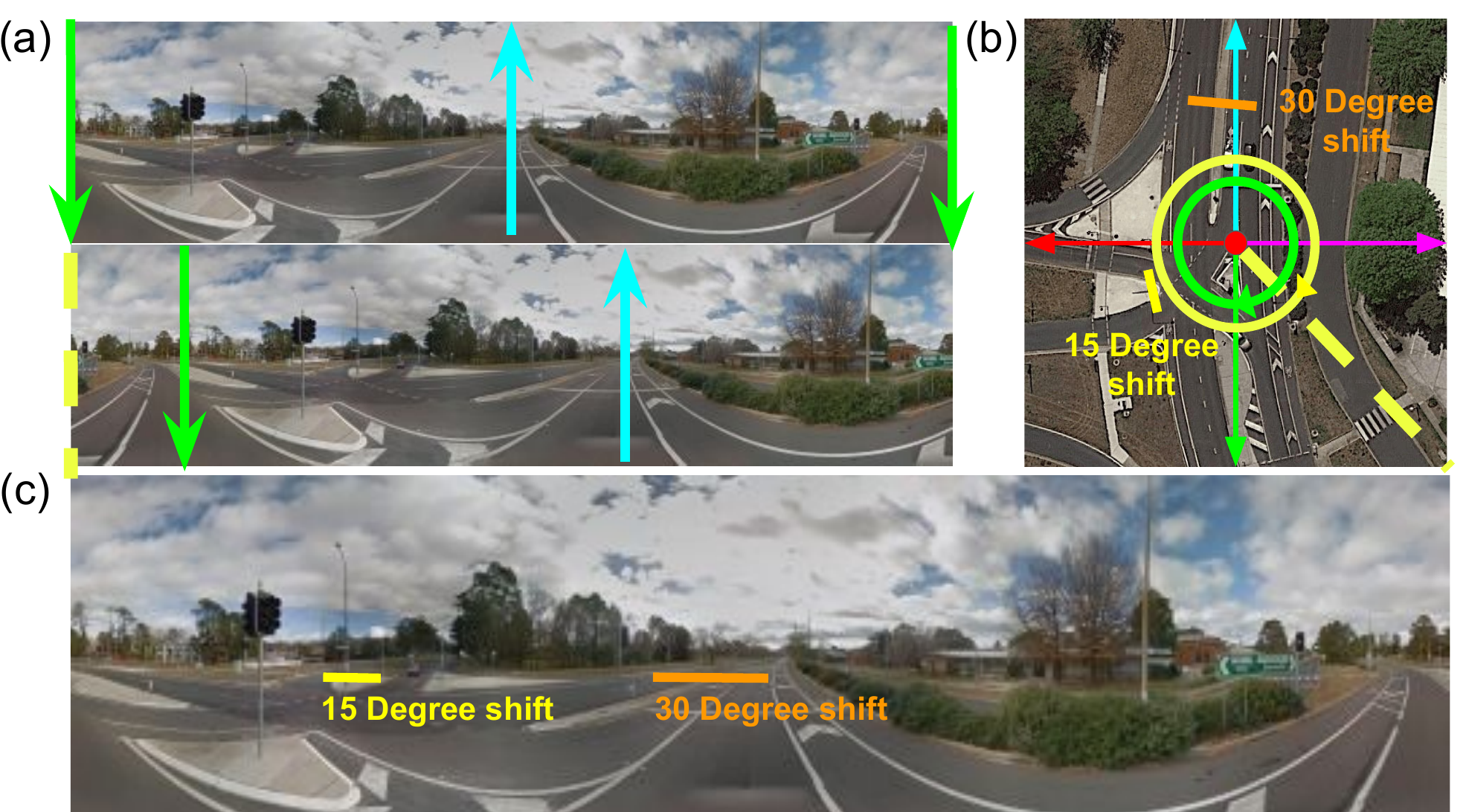}
  \vspace{-3mm}
  \caption{(a) Street-view image queries with 0$^{\circ}$ and 315$^{\circ}$ clockwise misalignments, (b) Geo-referenced satellite imagery as the reference for cross-view matching, and (c) Visualization of angle misalignment on a sample street-view image.}
  \Description{visualization of samples images}
  \label{fig:angle_definition}
  
  \vspace{-6mm}
\end{figure}
\section{Introduction}
Photos not only contain memories, but also provide us a brand new way to learn and perceive the world through others' eyes, to find details that one may have overlooked earlier, and to share emotions and knowledge with the community. With advances in hardware, personal high-quality cameras have become much more affordable. Many creators are keen on sharing photos on the internet. The captured images may not be as carefully calibrated as if they were taken by a dedicated multi-sensor system (\eg Google Street View vehicles), but the sheer volume of crowdsourced images surely provides rich information. If we can efficiently estimate the missing meta information (\eg geo-location, camera orientation) of those images and calibrate them for ``ready-to-use'' status, this enormous hidden treasure can help on various downstream tasks, \eg map information extraction~\cite{BalaliEtAl2015, MamidalaEtAl2019, ZamirEtAl2011}, car navigation and tracking \cite{HuEtAl2018, SenletElgammal2011, AgarwalEtAl2015, LeiEtAl2020, RegmiShah2021}, UAV positioning \cite{MajdikEtAl2015, LinEtAl2015}, hazard detection~\cite{WangEtAl2021a, ManiatEtAl2021, ChacraZelek2017, LeiEtAl2020}, social studies \cite{WangVermeulen2021}. To accomplish this goal, we must carry out three tasks: (a) adjusting the image upright~\cite{JungEtAl2017}, (b) finding the location, and (c) estimating the viewing angle of the camera.

Image-based geo-localization is a line of study aiming at inferring the camera location of street-view images. Among various geo-localization approaches, cross-view geo-localization uses geo-referenced aerial images (mostly satellite imagery). Given a street-view query image (\autoref{fig:angle_definition}(a)), the system finds the most similar match in a pool of satellite images (\autoref{fig:angle_definition}(b)), and then takes the satellite image center as the localization result. Thanks to its image retrieval nature, cross-view matching with satellite imagery can be applied in large-scale searches and gives promising results.

However, as a full decade has passed, finding only the location no longer satisfies the requirement to explore the full potential of street-view images. We encourage the community to embrace a more challenging task of \emph{estimating the fine-grained camera orientation of street-view images} to extend the overall objectives for three reasons:

\noindent 1) With the accurate camera orientation, information extracted from street-view images, especially from using single-image algorithms (\eg depth estimation, object detection), enables a wider range of real-world applications, \eg map creation, augmented reality, navigation. In practice, even a very small misalignment in orientation can propagate a large shift to the physical position of objects detected in images. For example, \autoref{fig:angle_definition} (c) shows that an orientation error of $15^{\circ}$ is large enough to mislocate an exit to another lane; an error of $30^{\circ}$ is sufficient to mistakenly assign attributes to the reverse direction of the road. 

\noindent 2) Nowadays, crowdsourced high-quality $360^{\circ}$ or wide-angle images can be taken by semi-professional 360$^\circ$ cameras or even by phones. These images are usually not carefully calibrated. It is more likely that the orientation information is missing but a rough location is labeled, rather than vice versa. Hence, the problem of finding the location of street-view images assuming the orientation is known is no longer realistic. 

\noindent 3) After surveying recent studies, we found that the performance of cross-view geo-localization is slowly reaching saturation. Based on our experiments, we observe that by introducing a finer granularity to orientation estimation, the performance of geo-localization can be further improved. We hypothesize that the saturation of performance is reached not because of the representation power of the networks, but a lack of precision in orientation to align two images. Finding fine-grained orientation could also potentially improve the performance of geo-localization of crowdsourced images.

Only a few studies have made progress or introduced new ideas on using the orientation characteristic~\cite{ZhuEtAl2020, ShiEtAl2020, LiuLi2019, CaiEtAl2019, VoHays2017, ZhaiEtAl2017}. Even fewer evaluates the orientation estimation result along with the geo-localization task ~\cite{VoHays2017, ZhaiEtAl2017, ZhuEtAl2020,ShiEtAl2020}. To the best of our knowledge, Shi~\etal~\cite{ShiEtAl2020} is the only work taking orientation prediction as the main problem to solve for large scale searching. However, we found the granularity of the estimation could be improved to better fit real-world use cases and the evaluation metrics do not give enough clarity. Hence, in this work, we formally define the problem and propose a set of metrics for evaluating the fine-grained estimation of camera orientation of street-view images. In summary, our contributions lie in the following aspects:
\vspace{-1mm}
\begin{itemize}[leftmargin=*]
    \item Define a south-aligned orientation alignment coordinate and a continuous absolute angle error coordinate for orientation estimation in cross-view matching with satellite imagery.
    \item Propose two methods to enhance the granularity of orientation estimation of street-view images without introducing any additional learnable parameters.
    \item Propose a set of metrics for orientation estimation, which is easier to understand and gives better clarity for real-world use-cases.
    \item Demonstrate that by incorporating fine-grained orientation estimation, the performance of geo-localization can be improved. 
    \vspace{-1mm}
\end{itemize}

\section{Fine-grained orientation estimation}
\begin{figure}
  
  \includegraphics[width=0.47 \textwidth, trim = 0cm 7.2cm 32.5cm 0cm, clip]{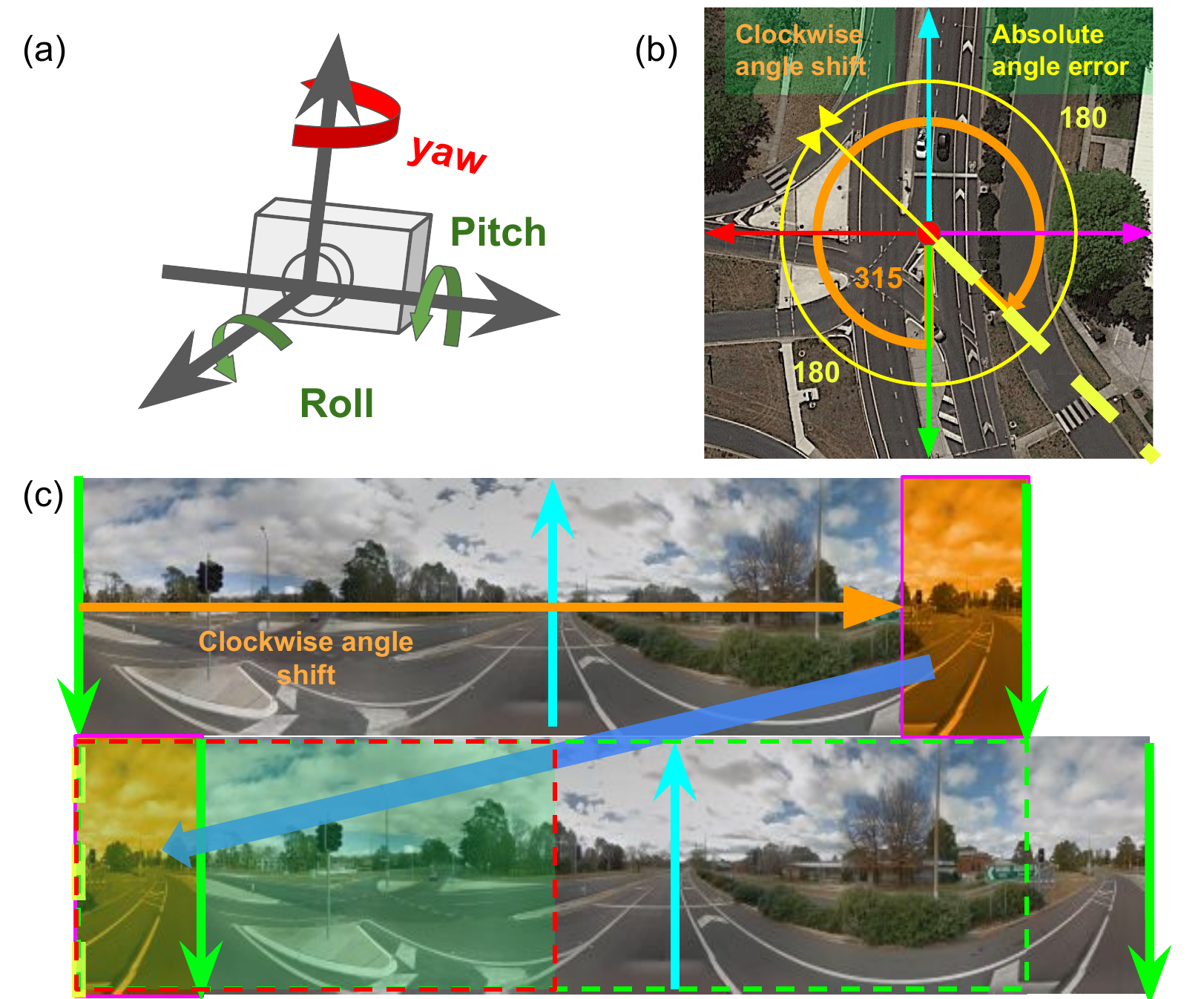}
  \caption{(a) Camera angles, (b) South-aligned reference coordinate for orientation shift, and (c) Visualization of rotation applied on images to create orientation misalignment.}
  \Description{visualization of reference system}
  \label{fig:angle_error}
  
  \vspace{-6mm}
\end{figure}

\subsection{Preliminaries}
Prior works on geo-localization~\cite{SenletElgammal2011, BansalEtAl2011a, LinEtAl2013a, WorkmanJacobs2015, AltwaijryEtAl2016, TianEtAl2017} leverage on different geo-referenced data sources, \eg GIS map images. In more recent years, \cite{VoHays2017} forms this task as an image retrieval problem that matches street-view image queries with geo-referenced satellite images and becomes one of the mainstream approaches for street-view imagery geo-localization. Given an upright street-view image $I_g$ (\autoref{fig:angle_error} (c) top) and a set of geo-referenced satellite imagery (\autoref{fig:angle_error} (b)), a system shall identify the satellite image $I_s$ at the same location as $I_g$ from a pool of satellite image candidates. The center location of $I_s$ is assigned to be the location of $I_g$.

Besides geo-localization, finding accurate camera orientation is the other critical task to prepare street-view images for ``ready-to-use'' status. \autoref{fig:angle_error} (a) shows the three camera angles required for calibration. The pitch and roll angle along with other camera distortion can be corrected by methods like \cite{JungEtAl2017}. After such corrections, the street-view images are upright and only left the yaw angle, the orientation, for estimation. In the commonly used datasets, the orientation of the street-view image is usually north-aligned, which means the center column of the image points to the Geographic North Pole (marked by the cyan arrow in \autoref{fig:angle_error} (c) top) to ensure it is aligned with the north direction of the geo-referenced satellite imagery (north marked by the cyan arrow in \autoref{fig:angle_error} (b)). It is commonly known that geo-localization obtains better performance if the orientations of the street-view images are known. However, only a few works on geo-localization \cite{VoHays2017, ShiEtAl2020, ZhuEtAl2020} propose new initiatives on orientation estimation. Among them, only \cite{ShiEtAl2020, ZhuEtAl2020} assign dedicated components to extract the orientation and use the estimated orientation explicitly in the geo-localization steps. We found each previous work \cite{VoHays2017, ShiEtAl2020, ZhuEtAl2020, LiuLi2019} has a slightly different definition of the orientation misalignment and error. Therefore, we provide a unified definition that could be beneficial for future works.

\subsection{Problem Formulation}
\subsubsection{Problem statement}

Given a set of upright street-view images $\mathcal{I}_g=\{I_g\}$ and a set of geo-referenced satellite images $\mathcal{I}_s=\{I_s\}$, which are paired and cropped at the same location of their paired street-view image, an orientation misalignment $\theta_\text{gt}$ is created for each street-view and satellite pair following the reference coordinate in section \ref{section:angle_shift_system}. For each street-view query image $I_g$, the similarity and orientation $\theta_\text{est}$ between the query $I_g$ and every satellite image candidate in $\mathcal{I}_s$ are estimated. The satellite candidates are ranked by their similarity. The center location of the top-1 satellite image and the estimated orientation with the correct match are extracted as the estimated location and orientation of the query image $I_g$. We aim to reduce the error between the estimated orientation $\theta_\text{est}$ and $\theta_\text{gt}$ (absolute error coordinate in section \ref{session:angle_error_system}), while maintaining or improving the recall of geo-localization.

\subsubsection{Reference coordinate for angle shift} \label{section:angle_shift_system}
To create the misalignment between the street-view image and geo-referenced satellite image, we adapt the augmentation method from \cite{ShiEtAl2020} to randomly shift the street-view images clockwise and record this angle shift $\theta_\text{gt}$ in a south-aligned reference coordinate. \autoref{fig:angle_error}(c) shows an example of shifting the street-view image $315^{\circ}$ clockwise. Semantically, the augmentation crops the right-most part outside the shifting angle (in the orange shade) and stitches it to the left-most column of the street-view image, and then crops the field of view (FOV) required for the output. In \autoref{fig:angle_error} (c), red box shows FOV of $180^{\circ}$, green box shows FOV of $360^{\circ}$. Pseudo-code of rotation and cropping is given in the supplementary materials. Although the original image pairs are north-aligned, we choose the south alignment as the reference, namely the alignment between the first column of the image to the south direction of the satellite image (marked in green arrows in \autoref{fig:angle_error}). This change is made for images with limited FOV. After cropping, the center column is shifted left-wards and is no longer aligned with the angle shift ground truth $\theta_\text{gt}$. For example, in a north-aligned coordinate, if a $360^{\circ}$ image is shifted by 90 degrees clockwise and cropped to FOV of $180^{\circ}$, the angle shift between the center column and the north direction becomes 45 degrees; if crop to $120^{\circ}$, the angle shift becomes 30 degrees. \autoref{tab:alignment method} shows different methods for creating misalignment. Compared to DSM~\cite{ShiEtAl2020}, our method is FOV invariant, $\theta_\text{gt}$ is consistent regardless of the FOV. Compared to \cite{ZhuEtAl2020}, rotating street-view image is easier to implement and our method avoids losing the corners of satellite images because of rotation and distortions caused by interpolation. 

\begin{table}[]
\small
\caption{Different methods to rotate and create misalignment.}
\vspace{-0.5em}
\label{tab:alignment method}
\begin{tabular}{|l|c|c|c|}
\hline
Methods & Rotation on & Alignment & FOV invariant \\ \hline
DSM \cite{ShiEtAl2020}     & Street-view & North     &               \\ \hline
Zhu \etal\cite{ZhuEtAl2020}     & Satellite   & South     & \checkmark              \\ \hline
Ours    & Street-view & South     &       \checkmark        \\ \hline
\end{tabular}
\vspace{-4mm}
\end{table}

\subsubsection{Reference coordinate for angle error} \label{session:angle_error_system}
We propose to calculate the absolute angle error between the estimated angle shift $\theta_\text{est}$ and ground truth angle shift $\theta_\text{gt}$. Centered to the $\theta_\text{gt}$, the errors count up to 180 degrees to the opposite direction (\autoref{fig:angle_error} (b)). Compared to using the $[0, 360]$ clockwise or $[-180, 180]$ angle error in \cite{VoHays2017, ZhaiEtAl2017, ZhuEtAl2020}, the absolute angle error coordinate has the following advantages:
\begin{itemize}[leftmargin=*]
    \item Avoid a sudden change of 360 degrees around the border case in $360^{\circ}$ clockwise system ($0 \leftrightarrow 360$) and $[-180, 180]$ system ($-180 \leftrightarrow 180$). Our absolute error coordinate is continuous everywhere, being more natural for defining the angle loss function.
    \item More fair to compute angle errors. For example, when the ground truth $\theta_\text{gt}=0^{\circ}$ and two estimations are $\theta_\text{est1}=170^{\circ}$ and $\theta_\text{est2}=240^{\circ}$. In a $360^{\circ}$ system, the angle error for $\theta_\text{est1}$ is $170^{\circ}$ and for $\theta_\text{est2}$ is $240^{\circ}$, although $\theta_\text{est2}$ is closer to $\theta_\text{gt}$ with an absolute error of $120^{\circ}$. The system shall favour  $\theta_\text{est2}$ instead of $\theta_\text{est1}$.
    \item Easier to calculate angle error from two angles shifts given in the south-aligned coordinate. Given $\theta_1$ and $\theta_2$ in south-aligned coordinate, their angle difference can be calculated as:
    \begin{equation}
        \theta_\text{diff} = 180^{\circ} - \left| |\theta_1 - \theta_2 | - 180^{\circ} \right|.
    \end{equation}
\end{itemize}

\subsection{Method}
\begin{figure}[!t]
  \includegraphics[width=0.47 \textwidth, trim = 2.6cm 0cm 12.9cm 0cm, clip]{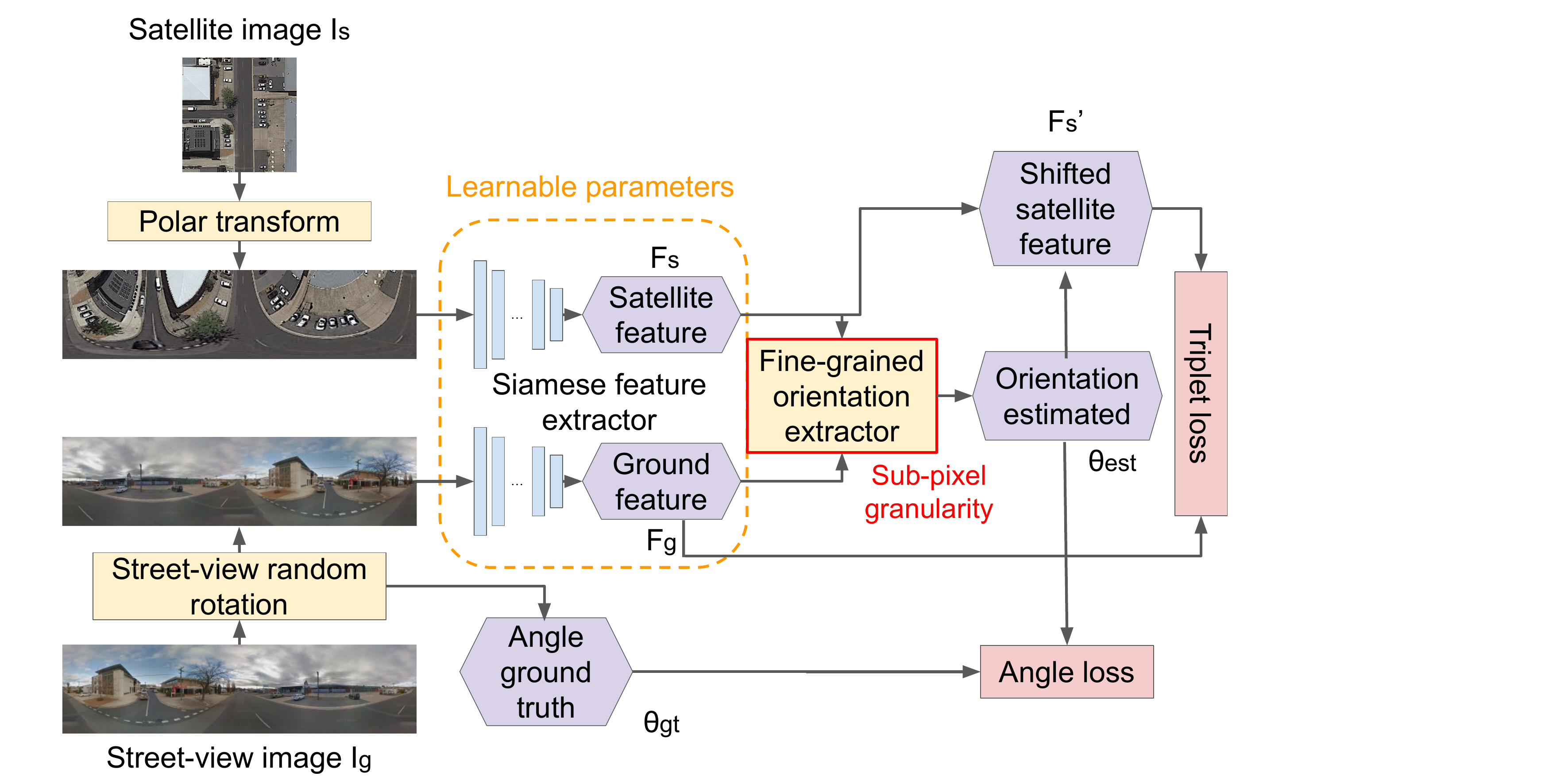}
  \vspace{-0.8em}
  \caption{Overall architecture for geo-localization and orientation estimation. Example shows with 360$^{\circ}$ FOV.}
  \Description{visualization of reference system}
  \label{fig:structure}
  \vspace{-1em}
\end{figure}
We adapt the architecture from DSM~\cite{ShiEtAl2020} and proposed two different approaches to obtain fine-grained orientation estimation without introducing any additional learnable parameters. On top of the commonly used triplet loss \cite{VoHays2017}, we defined a new angle loss based on the absolute angle error system (see Section \ref{session:angle_error_system}) to learn the angle shift explicitly.
\subsubsection{\hspace{-0.7em} Overview}
\autoref{fig:structure} shows the overall architecture of our geo-localization and orientation estimation system. 

\noindent\textbf{Pre-processing:} Before passing the input images to the feature extractors, the following pre-processes are applied to street-view images and satellite images respectively:
\begin{itemize}[leftmargin=*]
    \item In training, street-view images are randomly rotated to introduce orientation misalignments, following section \ref{section:angle_shift_system}. The ground truth shift in feature space $w_\text{gt}$ is recorded to calculate the angle loss and orientation estimation accuracy, where $\theta_\text{gt} = w_\text{gt} / \text{width}(F_s) *360^\circ$ and $F_s$ is the extracted feature from satellite images.
    \item Satellite images are \emph{polar-transformed} \cite{ShiEtAl2020} to a similar viewing point and size as the street-view images to physically reduce the gap between street-view and satellite view images. \autoref{fig:structure} left top shows the effect of polar transformation.

\end{itemize}

\noindent\textbf{Feature extraction:} The pre-processed satellite and street-view images are sent to \emph{Siamese feature extractors} (in the middle of \autoref{fig:structure}), which are modified based on VGG16~\cite{SimonyanZisserman2015}. Following \cite{ShiEtAl2020}, the extractors consist of the first 10 layers of VGG16 and 3 additional convolutional layers to further compress the feature maps on the vertical axis. The last three layers have an output feature depth of [256, 64, 16] and stride [(2, 1), (2, 1), (1, 1)] in the vertical and horizontal axes. For full $360^{\circ}$ inputs with size [128, 512, 3], two feature maps $F_s$ and $F_g$ with a size of [4, 64, 16] in [H, W, C] are extracted respectively. The two branches have the same architecture without weight sharing. 

\noindent \textbf{Fine-grained orientation estimation:} The extracted features $F_s$ and $F_g$ are processed by the fine-grained orientation extractor to infer the sub-pixel level angle shift $\theta_\text{est}$. Then the satellite features $F_s$ are shifted and cropped ($F_s'$) to be aligned with the street-view features $F_g$ in orientation and FOV. The output features $F_s'$ and $F_g$ are used to calculate the triplet loss. Moreover, an angle loss is used to provide direct supervision on the orientation estimation. During inference, if the orientation between satellite and street-view is known ($\theta_\text{gt}=0$), the orientation extractor is not used, only cropping of $F_s$ to have the same FOV is applied; if the orientation is unknown, the fine-grained orientation extractor finds the alignment. Shifting and cropping of $F_s$ are applied. Given a query street-view image, the feature similarities between the query and all possible satellite candidates are calculated along with the orientation estimation.

\subsubsection{Fine-grained orientation extractor}
To find fine-grained orientation (at a sub-degree level), we propose two methods to increase the granularity of the estimation without increasing the number of learnable parameters, compared to DSM \cite{ShiEtAl2020}. After obtaining the features $F_s$ and $F_g$, the cross-correlation is calculated as:
\begin{equation}
\label{eq:correlation}
    (F_g \star F_s)[w] = \sum\nolimits_{m=1}^{W_g} F_g [m] F_s [(m+w)_{\text{mod} \: {W_s}}], 
\end{equation}
where $F_{g/s} [m]$ is a slice of features at horizontal position $m$ across all height and channels. $W_s$ and $W_g$ are the width of $F_s$ and $F_g$. The position with the highest cross-correlation value is taken as the estimated angle shift $w_\text{est}$ in the feature space. As our input images have a width of 512 pixels, the shifting unit is about $0.7$ degree ($360^\circ/512$ pixels). However, $F_s$ and the cross-correlation result (\autoref{eq:correlation}) only have a width of 64 pixels, which makes the orientation extractor have a maximum resolving power of $5.625$ degree ($360^\circ/64$ pixels). The granularity of the result surely cannot recover the actual angle shift for our study and is also not enough for real-world applications. To refine the granularity of the estimation, we proposed two approaches.

\noindent \textbf{Feature interpolation (FI):} Following \autoref{alg:FI}, both $F_s$ and $F_g$ are interpolated with a scaling factor $S$ before calculating the fine-grained cross-correlation curve. In our implementation, we increase the granularity by 10 times. The bin number with the maximum cross-correlation curve is extracted and is divided by $S$ to obtain sub-pixel level estimation $w_\text{est}$. The resolving power of the model is refined to $0.5625$ degree ($360^\circ/640$ pixels).

\begin{algorithm}[t]
\caption{Feature Interpolation (FI)}
\label{alg:FI}
\small
\KwIn{street-view features $F_g$, satellite features $F_s$, scaling factor $S$.}
\KwOut{Estimated orientation in feature space  $w_\text{est}$}

$F_g^s = \text{interpolate}(F_g, S), \quad F_s^s = \text{interpolate}(F_s, S)$

$(F_g^s \star F_s^s)[w] = \text{cross-corr}(F_g^s, F_s^s)$ \hfill // eqn (2)

$w^s = \max_\text{position}((F_g^s \star F_s^s)[w])$ \hfill // find shift with max value

$w_\text{est} = w^s / S$ \hfill // get sub-pixel position

\Return{$w_\text{est}$}
\end{algorithm}

\noindent \textbf{Curve smoothing (CS):} Following \autoref{alg:CS}, the cross-correlation curve $(F_g \star F_s)[w]$ is calculated with the original resolution (64 bins). To smooth the curve with scaling factor $S=10$, the coarse cross-correlation curve is transformed to frequency domain with Fast Fourier Transform (FFT) and zero-padded of $(S-1)$ times to the middle of the curve:
\begin{displaymath}
\begin{split}
    \text{ZP}(F[w], S) = \small\text{Concat}(&F[0 \colon \!\small\text{int}(W/2)], \\
    &\small\text{zeros}[0 \colon \! (S\!-\!1)W], F[\small\text{int}(W/2)+1\! \colon\!]),
\end{split}
\end{displaymath}
where $W$ is the width of the $F[w]$. The output of zero-padding is again converted back by Inverse FFT (IFFT). The fine-grained orientation extractor with CS has a resolving power of 0.5625 degree. 

Both CS and FI provide the flexibility to adjust the granularity of the orientation extraction via the changeable scaling factor.

\begin{algorithm}[t]
\caption{Curve Smoothing (CS)}
\label{alg:CS}
\small
\KwIn{street-view features $F_g$, satellite features $F_s$, scaling factor $S$.}
\KwOut{Estimated orientation in feature space  $w_\text{est}$}

$(F_g \star F_s)[w] = \text{cross-corr}(F_g, F_s)$ \hfill // coarse curve

$((F_g \star F_s)[w])_\text{fft-padded} =\text{zero-padding}(\text{\small FFT}((F_g \star F_s)[w]),\! S)$

$((F_g \star F_s)[w])_\text{smooth} = \text{IFFT}(((F_g \star F_s)[w])_\text{fft-padded})$

$w^s = \max_\text{position}(((F_g \star F_s)[w])_\text{smooth})$

$w_\text{est} = w^s / S$ \hfill // sub-pixel position

\Return{$w_\text{est}$}

\end{algorithm}

\subsubsection{Objective functions with angle loss}
To have the direct supervision on angle estimation, we propose an angle loss based on the absolute angle error (section \ref{session:angle_error_system}). Given the ground truth orientation in feature space $w_\text{gt}$, the estimated orientation $w_\text{est}$ and the width of the feature map space $W$, the angle loss is given as:

\begin{equation}
    L_{\small\text{angle}} = (0.5W - \left| |w_\text{gt} - w_\text{est} | - 0.5W \right|)/0.5W,
\end{equation}
which is equivalent to the rate of the angle error to the maximum error of 180$^\circ$. Note that this loss is only applied to matched pairs.

For the geo-localization task, we utilize the commonly used weighted soft-margin triplet loss \cite{HuEtAl2018, ShiEtAl2020, ShiEtAl2019, ShiEtAl2019a}. Given a triplet consists of an anchor query image $A$ in one view, the positive sample $P$ (correct match) and a negative sample $N$ in the other view, the feature ($F_A$) extracted from $A$  shall have a smaller distance to the shifted and cropped feature ($F_P^\prime$) from the $P$ than the shifted and cropped feature ($F_N^\prime$) from $N$. We take the cosine distance between features in our implementation. The loss function is given as:
\begin{equation}
\begin{gathered}
    L_{\small\text{match}} = \ln(1 + \exp(\alpha(D(F_A,F_P^\prime)-D(F_A,F_N^\prime)))) \\
    D(F_1, F_2) = 2(1-\text{cos}(F_1, F_2)),
\end{gathered}
\end{equation}
where the $\alpha=10$, following \cite{ShiEtAl2020}. For a batch size of $B$, each query can form $(B-1)$ triplets. In each matching direction (street $\rightarrow$ satellite or satellite $\rightarrow$ street), $B(B-1)$ triplets are constructed. We enforce the matching on both directions and have totally $2B(B-1)$ triplets in each mini-batch. The overall objective function is:
\begin{equation}
    L = L_{\small\text{match}}(A, P, N) + \beta \: L_{\small\text{angle}}(A, P).
\end{equation}

\subsection{Metrics} \label{section: metrics}
Few metrics have been proposed to evaluate the accuracy of orientation estimation. \cite{VoHays2017, ZhuEtAl2020, ZhaiEtAl2017} visualized a histogram of estimation error in the range $[-180^\circ, 180^\circ]$, with 
granularity around 7 to 10 degrees. \cite{ZhuEtAl2020} gives the percentage of images with estimation error within $[-3.5^\circ, 3.5^\circ]$. \cite{ShiEtAl2020} defines two metrics: 1) the accuracy of the estimations within an error of $\pm10\%$ of FOV. 2) median of the angle errors of images that are correctly geo-localized (recall@top1). Median is used instead of mean because of some images are estimated to the opposite direction with an error of $180^{\circ}$. This large error has a huge impact on the mean error, but may not have a large impact on geo-localization. To evaluate the angle error independently from the localization and to provide more clarity on fine-grained orientation estimation for downstream tasks, we propose the following orientation-centric metrics.

\subsubsection{Fine-grained histogram} A histogram $H(\theta)$ at $1^{\circ}$ granularity is calculated for the absolute angle errors. For every image in the test set, given the ground truth $\theta_\text{gt}$ and the estimation $\theta_\text{est}$:
\begin{equation}
\label{eq:histogram}
\begin{split}
    \theta_\text{err} &= 180^{\circ} - \left| |\theta_\text{gt} - \theta_\text{est} | - 180^{\circ} \right|,\\
    H(\theta_i) &= H(\theta_i) + 1, \quad \text{where}  \: \theta_{i} -1 \leq \theta_\text{err} <\theta_{i}.
\end{split}    
\end{equation}
With the fine-grained histogram, 1-to-1 visual comparison between models and the accumulated accuracy curve at any specific degree can be retrieved easily. It shows the distribution and reliability of the orientation estimation, which are crucial for downstream tasks.  

\subsubsection{Mean angle error} 
The mean of angle errors of all test images is calculated to evaluate the orientation performance independently on geo-localization for two reasons: 1) There exist alternative sources to obtain location-tag, \eg social media, and the orientation estimation remains to be the bottleneck for downstream tasks. Any missing orientation information gives the testing images a $180^{\circ}$ uncertainty. 2) A $180^{\circ}$ error may not affect the geo-localization result, but could be the worse case for many downstream tasks, \eg navigation. Hence, the mean is used to calculate the error linearly for the entire test set.

\subsubsection{Accuracy below specific threshold} The accuracy of test images with an estimation error below a specific threshold is calculated. For a given fine-grained histogram $H(\theta)$ (\autoref{eq:histogram}), the rate below $x^\circ$ is given as:
\begin{equation}
    \text{r@x}^{\circ} =  \sum_{i=1}^{x} H(\theta_i)/\sum_{i=1}^{180} H(\theta_i).
\end{equation}
With these metrics, the users can decide whether the estimated orientation fits the downstream tasks, which usually come with a tolerance of orientation errors.

\section{Experimental settings}

\subsection{Datasets and Metrics}
We evaluate the orientation estimation and geo-localization on two commonly used datasets, CVUSA~\cite{WorkmanEtAl2015} and CVACT~\cite{LiuLi2019}. Both datasets contain 35,532 training street-satellite matched pairs and 8,884 test pairs. All images are angle-aligned. At the training time, random shifting and cropping (if with limited FOV) are applied to street-view images. In testing, we followed the orientation shift given to each matched pair in \cite{ShiEtAl2020}. Note that the two datasets are collected in the US and Australia separately and have a non-negligible domain shift. CVUSA contains a mix of commercial, residential, suburban, and rural areas and CVACT leans towards urban/suburban styles. Additionally, the satellite images in CVUSA have a higher ground coverage but a lower resolution than CVACT, which introduces another substantial domain shift. All street-view images and polar-transformed satellite images are resized to $[128, 512]$ pixels in height and width. To evaluate the performance of orientation estimation, mean angle error, $r@2^{\circ}$, $r@5^{\circ}$ are used. Visualization of angle error histogram is presented to compare the differences among models. To demonstrate the improvement in geo-localization brought by fine-grained orientation estimation, recall at the top 1 sample (r@1) is used for known/unknown orientation test. Additionally, model generalization test results in across and mixed dataset setting are shown in the supplementary materials.

\subsection{Implementation Details}
All our models and our re-implementation of DSM~\cite{ShiEtAl2020}$^*$ are trained with unknown orientation. The first 10 layers of the VGG16 based feature extractors use the pre-trained weights on ImageNet~\cite{DengEtAl2009} and the last three layers are initialized randomly. Note that in our model all parameters are learnable, while the first 10 layers of DSM are frozen. For fair comparison, batch size B is set to 32 as \cite{ShiEtAl2019a, ShiEtAl2020,YangEtAl2021c,WangEtAl2021,ZhangEtAl2022}. We use Adam~\cite{KingmaBa2017} optimizer with an initial learning rate of $11e^{-5}$, learning rate decrease on plateau is applied with factor 0.5, patience 8. The maximum training time is set to 200 epochs and the early stopping threshold is 30 epochs. For unknown orientation, images can rotate up to 360$^\circ$. The rotation unit is about 0.70$^\circ$, corresponding to the shift in one pixel of the input image. The weight for angle loss $\beta$=0.3. The average results of three instances are reported in the following sections, if not otherwise specified.

\section{Results}

\subsection{Fine-grained Orientation Estimation}

\begin{figure}
  \includegraphics[width=0.47 \textwidth, trim = 0.4cm 20.4cm 42.2cm 0cm, clip]{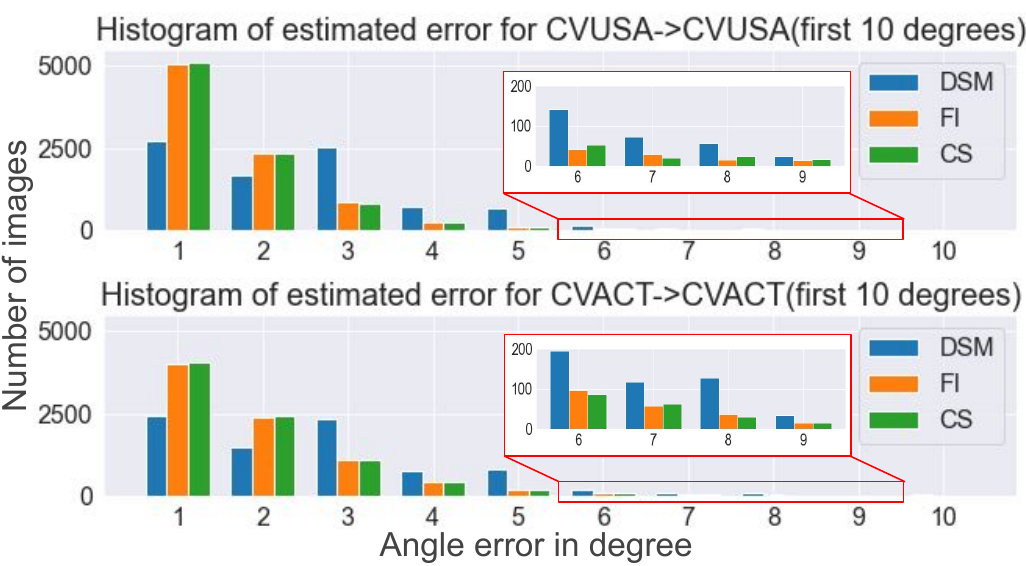}
  \vspace{-4mm}
  \caption{Histogram of angle errors for the best instance of each model on 360$^\circ$ image test. Each set has 8,884 images.}
  \label{fig:angle_hist}
\end{figure}

\begin{table}[]
\small
\caption{Orientation estimation on two datasets. The baseline DSM\cite{ShiEtAl2020}$^*$ is used to compute the gain in the brackets.}
\label{tab:angle error}
\vspace{-4mm}
\begin{tabular}{|c|ccc|ccc|}
\hline
Model     & \multicolumn{1}{c|}{\begin{tabular}[c]{@{}c@{}}Mean\\ error $\downarrow$ \end{tabular}}           & \multicolumn{1}{c|}{\begin{tabular}[c]{@{}c@{}}r@2$^\circ$\\ (\%)\end{tabular}}               & \begin{tabular}[c]{@{}c@{}}r@5$^\circ$\\ (\%)\end{tabular}              & \multicolumn{1}{c|}{\begin{tabular}[c]{@{}c@{}}Mean\\ error $\downarrow$ \end{tabular}}            & \multicolumn{1}{c|}{\begin{tabular}[c]{@{}c@{}}r@2$^\circ$\\ (\%)\end{tabular}}               & \begin{tabular}[c]{@{}c@{}}r@5$^\circ$\\ (\%)\end{tabular}              \\ \hline
                                                         & \multicolumn{3}{c|}{CVUSA}                                                                                                                                                                                                                    & \multicolumn{3}{c|}{CVACT}                                                                                                                                                                                                                    \\ \hline
Zhai~\cite{ZhaiEtAl2017}                                                      & \multicolumn{1}{c|}{-}                                                              & \multicolumn{1}{c|}{$\le$15}                                                    & -                                                               & \multicolumn{1}{c|}{-}                                                              & \multicolumn{1}{c|}{-}                                                                & -                                                               \\ \hline
Zhu~\cite{ZhuEtAl2020}                                                      & \multicolumn{1}{c|}{-}                                                              & \multicolumn{1}{c|}{$\le$24}                                                    & -                                                               & \multicolumn{1}{c|}{-}                                                              & \multicolumn{1}{c|}{-}                                                                & -                                                               \\ \hline
DSM~\cite{ShiEtAl2020}*                                                      & \multicolumn{1}{c|}{5.29$^\circ$}                                                           & \multicolumn{1}{c|}{47.49}                                                            & 93.25                                                           & \multicolumn{1}{c|}{6.26$^\circ$}                                                           & \multicolumn{1}{c|}{44.13}                                                            & 88.31                                                           \\ \hline
\begin{tabular}[c]{@{}c@{}}Ours (FI)\\ Gain\end{tabular} & \multicolumn{1}{c|}{\begin{tabular}[c]{@{}c@{}}3.78$^\circ$\\ (1.51$^\circ$)\end{tabular}}          & \multicolumn{1}{c|}{\begin{tabular}[c]{@{}c@{}}82.13\\ (34.64)\end{tabular}}          & \textbf{\begin{tabular}[c]{@{}c@{}}96.77\\ (3.52)\end{tabular}} & \multicolumn{1}{c|}{\begin{tabular}[c]{@{}c@{}}4.88$^\circ$\\ (1.38$^\circ$)\end{tabular}}          & \multicolumn{1}{c|}{\begin{tabular}[c]{@{}c@{}}70.87\\ (26.74)\end{tabular}}          & \begin{tabular}[c]{@{}c@{}}92.29\\ (3.98)\end{tabular}          \\ \hline
\begin{tabular}[c]{@{}c@{}}Ours (CS)\\ Gain\end{tabular} & \multicolumn{1}{c|}{\textbf{\begin{tabular}[c]{@{}c@{}}3.77$^\circ$\\ (1.52$^\circ$)\end{tabular}}} & \multicolumn{1}{c|}{\textbf{\begin{tabular}[c]{@{}c@{}}82.42\\ (34.93)\end{tabular}}} & \begin{tabular}[c]{@{}c@{}}96.75\\ (3.50)\end{tabular}           & \multicolumn{1}{c|}{\textbf{\begin{tabular}[c]{@{}c@{}}4.75$^\circ$\\ (1.51$^\circ$)\end{tabular}}} & \multicolumn{1}{c|}{\textbf{\begin{tabular}[c]{@{}c@{}}72.28\\ (28.15)\end{tabular}}} & \textbf{\begin{tabular}[c]{@{}c@{}}92.43\\ (4.12)\end{tabular}} \\ \hline
\end{tabular}
\vspace{-1em}
\end{table}

\autoref{tab:angle error} shows the performance of our models on orientation estimation on 360$^\circ$ images along with previous studies \cite{ShiEtAl2020, ZhaiEtAl2017, ZhuEtAl2020}. As only a few works report their results on orientation estimation and use different metrics, we re-implement DSM~\cite{ShiEtAl2020} to conduct a full comparison using the newly proposed metrics. The re-implemented DSM~\cite{ShiEtAl2020}$^*$ is used as the baseline to calculate the improvement shown in the bracket. Both our models, FI and CS, show significant improvement in the mean error, r@2$^\circ$ and r@5$^\circ$ for all test cases. 

Both datasets have a similar mean error improvement of about 1.38$^\circ$ to 1.52$^\circ$ from their original mean error of 5.29$^\circ$ and 6.26$^\circ$. However, CVUSA observed higher absolute improvement on r@2$^\circ$ (about 35\%) than CVACT (about 28\%). Comparing the histogram visualized in \autoref{fig:angle_hist}, the error distribution of CVUSA is further pushed to the lower angle error region than the distribution of the CVACT results. Around 57\% of the test dataset of CVUSA obtained an orientation estimation with an error below 1$^\circ$, which is around 45\% for CVACT. For CVACT, some test cases that are not pushed below the 2$^\circ$ error region are still successfully reduced within 5$^\circ$. We hypothesize that the CVUSA dataset contains a larger portion of suburban, rural areas than CVACT, which leads the images in CVUSA naturally to have less obvious features, such as buildings, to leverage. This makes the precise orientation estimation has more influence on CVUSA than CVACT.

Between FI and CS, FI interpolates the coarse feature maps with a large scaling number to generate a fine-grained correlation curve with new values, while CS obtains sub-pixel correlation curve values by smoothing the original curve. From our experiment results in \autoref{tab:angle error}, CS obtains slightly better results than FI on orientation extraction. However, FI gives a more fundamental fine-grained orientation curve generation. It could be useful when prior knowledge of the rough orientation is available, which can be added to the street-view features before the orientation curve is generated.

\begin{table}[]
\small
\tabcolsep=2.55mm
\caption{Evaluation on geo-localization for the two datasets. The best results of FI, CS and DSM~\cite{ShiEtAl2020}* are reported.}
\label{tab:geo-localization}
\begin{tabular}{|c|cc|cc|}
\hline
Model                                                         & \multicolumn{1}{c|}{\begin{tabular}[c]{@{}c@{}}known\\ r@1 (\%)\end{tabular}} & \begin{tabular}[c]{@{}c@{}}unknown\\ r@1 (\%)\end{tabular} & \multicolumn{1}{c|}{\begin{tabular}[c]{@{}c@{}}known\\ r@1 (\%)\end{tabular}} & \begin{tabular}[c]{@{}c@{}}unknown\\ r@1 (\%)\end{tabular} \\ \hline
\multicolumn{1}{|l|}{}                                        & \multicolumn{2}{c|}{CVUSA}                                                                                                                 & \multicolumn{2}{c|}{CVACT}                                                                                                                 \\ \hline
CVM-NET~\cite{HuEtAl2018}                                                      & \multicolumn{1}{c|}{22.47}                                                    & -                                                          & \multicolumn{1}{c|}{20.15}                                                        & -                                                          \\ \hline
Liu \& Li~\cite{LiuLi2019}                                                      & \multicolumn{1}{c|}{40.79}                                                    & -                                                          & \multicolumn{1}{c|}{46.96}                                                        & -                                                          \\ \hline
CVFT~\cite{ShiEtAl2019}                                                          & \multicolumn{1}{c|}{61.43}                                                    & -                                                          & \multicolumn{1}{c|}{61.05}                                                    & -                                                          \\ \hline
SAFA~\cite{ShiEtAl2019a}                                                           & \multicolumn{1}{c|}{89.84}                                                    & -                                                          & \multicolumn{1}{c|}{81.03}                                                    & -                                                          \\ \hline
LPN-SAFA~\cite{WangEtAl2021}                                                      & \multicolumn{1}{c|}{92.83}                                                    & -                                                          & \multicolumn{1}{c|}{83.66}                                                    & -                                                          \\ \hline
DSM~\cite{ShiEtAl2020}$^*$                                                           & \multicolumn{1}{c|}{93.57}                                                    & 80.75                                                      & \multicolumn{1}{c|}{83.88}                                                    & 75.24                                                      \\ \hline
L2LTR~\cite{YangEtAl2021c}                                                          & \multicolumn{1}{c|}{94.05}                                                    & -                                                          & \multicolumn{1}{c|}{84.89}                                                    & -                                                          \\ \hline
GAN-SAFA~\cite{TokerEtAl2021}                                                      & \multicolumn{1}{c|}{92.56}                                                    & -                                                          & \multicolumn{1}{c|}{83.28}                                                    & -                                                          \\ \hline
SSANET~\cite{ZhangEtAl2022}                                                        & \multicolumn{1}{c|}{91.52}                                                    & -                                                          & \multicolumn{1}{c|}{84.23}                                                    & -                                                          \\ \hline
\begin{tabular}[c]{@{}c@{}}SEH VGG16\\ bs 30~\cite{GuoEtAl2022a}\end{tabular}     & \multicolumn{1}{c|}{94.46}                                                    & -                                                          & \multicolumn{1}{c|}{-}                                                        & -                                                          \\ \hline
\begin{tabular}[c]{@{}c@{}}SEH VGG16\\ bs 120~\cite{GuoEtAl2022a}\end{tabular}    & \multicolumn{1}{c|}{95.11}                                                    & 85.36                                                      & \multicolumn{1}{c|}{84.75}                                                    & 78.13                                                      \\ \hline
\begin{tabular}[c]{@{}c@{}}SEH FCANet18\\ bs 120~\cite{GuoEtAl2022a}\end{tabular} & \multicolumn{1}{c|}{95.04}                                                    & 85.37                                                      & \multicolumn{1}{c|}{85.13}                                                    & 77.41                                                      \\ \hline
TransGeo~\cite{ZhuEtAl2022}                                                      & \multicolumn{1}{c|}{94.08}                                                    & -                                                          & \multicolumn{1}{c|}{84.95}                                                    & -                                                          \\ \hhline{|=|=|=|=|=|}
Ours (FI)                                                     & \multicolumn{1}{c|}{\textbf{95.50}}                                           & \textbf{85.45}                                             & \multicolumn{1}{c|}{\textbf{86.79}}                                           & 80.41                                                      \\ \hline
Ours (CS)                                                     & \multicolumn{1}{c|}{95.37}                                                    & 85.39                                                      & \multicolumn{1}{c|}{86.22}                                                    & \textbf{80.79}                                             \\ \hline
\end{tabular}
\vspace{-4mm}
\end{table}

\subsection{Geo-localization}
Finding fine-grained orientation does not only provide additional orientation information but also improves the performance of geo-localization. We evaluate the geo-localization result of our models in known/unknown orientation tests. The r@1 of CVUSA and CVACT are shown in \autoref{tab:geo-localization}. Note that the performance of the best instance of our  FI, CS  and DSM~\cite{ShiEtAl2020}$^*$ are reported to have a fair comparison. Compared to DSM~\cite{ShiEtAl2020}$^*$, r@1 for known/unknown orientation test on CVUSA improved by 1.93\%, 4.70\% for FI and 1.80\%, 4.64\% for CS; on CVACT are improved by 2.91\%,	5.17\% for FI and 2.34\%, 5.55\% for CS. Additional results on across dataset and mixed dataset tests and visualization of the top 5 best matched and worst mismatched cases are shown in the supplementary materials.

We achieves better r@1 than all existing methods; especially on CVACT obtaining absolute improvement of 1.66\% and 2.66\% on known and unknown orientation test, without implementing additional sampling strategy or computational expensive architecture. Note that SEH~\cite{GuoEtAl2022a} and our models have the same origin from DSM, its work on sampling strategies is in another disentangled research direction. The two methods do not contradict each other and can be applied together with proper adaptation.

\subsection{Ablation Study}

\subsubsection{Location matched vs all test images}
\begin{figure}[!t]
  \vspace{-1em}
  \includegraphics[width=0.47 \textwidth, trim = 0.4cm 24.2cm 42.3cm 0cm, clip]{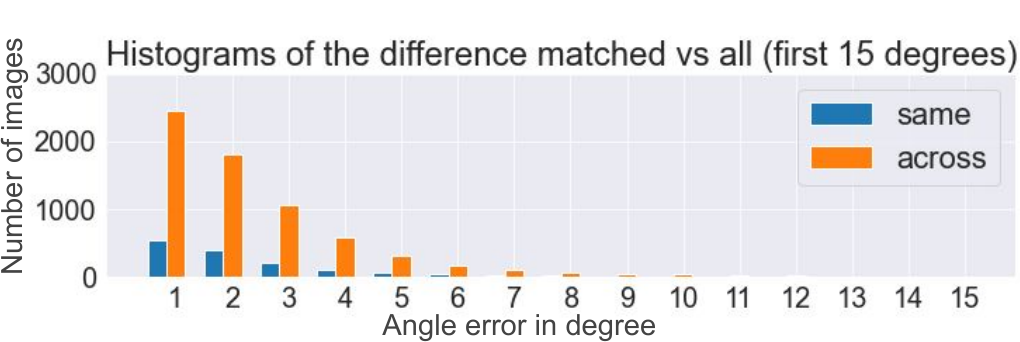}
  \vspace{-1em}
  \caption{Histogram of the difference between matched cases and all cases for CVACT same/across dataset tests. Most of the removed results are with good orientation estimation.}
  \Description{visualization of samples images}
  \label{fig:cvact_diff}
  \vspace{-1em}
\end{figure}

DSM~\cite{ShiEtAl2020} analyze the orientation estimation on the test images that are corrected geo-localized, while \cite{ZhaiEtAl2017,VoHays2017} evaluates the orientation estimation independently from geo-localization. Both methods have valid assumptions, hence we compare the results when only the location-matched images or all test images are evaluated. On top of testing the trained models on their own test set, \autoref{tab:matched vs all}  also includes the performance in across dataset tests, namely tests on the images from another domain, to demonstrate the difference in results by using the two approaches. In \autoref{tab:matched vs all}, the best instance of FI CVACT and CS CVUSA is shown, `all' means the results of all test images; `matched' means the results of only the matched images; `matched to all' means the results of the matched cases divided by the number of images of the full test set (8,884 images).  After removing unmatched cases, the mean error drops drastically, especially for across dataset tests. The r@2$^\circ$ also increases for all tests. Both indicate the images with good geo-localization results have a better orientation estimation in general. However, if consider orientation estimation as an individual problem, any missing estimation given the test images a 180$^\circ$ uncertainty, filtering results based on the location correctness can end up having a very low percentage of correctly estimated images in the full dataset. For across dataset tests, r@2$^\circ$ (matched to all) drops to 9.02\% and 12.69\%, although the models have the ability to give high-quality orientation to 57.09\% and 55.94\% of the full test images. \autoref{fig:cvact_diff} shows the majority of the removed cases obtain a high to medium quality orientation estimation. Mislocated images are not necessarily having low-quality orientation estimation. Additionally, evaluating only on location-matched images can lead to an unfair comparison, \eg a model can trick the evaluation by having only one correctly located image with a perfect estimated orientation. Hence, we believe the evaluation of orientation estimation shall be independent of geo-localization, unless it is for specific use cases.   

\subsubsection{Fixed parameters, angle loss, fine-grained}
The parameters in the first 10 layers of the baseline model DSM~\cite{ShiEtAl2020} are fixed. For a fair comparison, \autoref{tab:nofix vs angle vs FI CS} shows the average performance for 5 different settings on both datasets. DSM~\cite{ShiEtAl2020}$^*$ indicate our re-implementation of DSM, DSM (all) refers to the model with all parameters unfixed. `AL' refers to angle loss, FI and CS refer to the two methods we proposed for fine-grained orientation estimation. After releasing all parameters, the performance in all metrics improved compared to baseline. Adding angle loss to DSM (all) alone did not bring many additional benefits as the granularity of the prediction is still at a coarse level. After increasing the granularity, models with `AL' +`FI'/`CS' show substantial improvements in orientation estimation and perform better on geo-localization for both datasets. The full result of different training settings and angle loss weight is given in the supplementary materials.

\begin{table}[]
\small
\caption{Orientation estimation evaluated on all test data or location matched data. Matched to all means the result evaluated on matched cases to the full dataset, 8,884 images.}
\label{tab:matched vs all}
\vspace{-2mm}
\begin{tabular}{|l|c|c|c|c|}
\hline
                          & \begin{tabular}[c]{@{}c@{}}CVACT-\textgreater \!\\ CVACT\end{tabular} & \begin{tabular}[c]{@{}c@{}}CVUSA-\textgreater \!\\ CVUSA\end{tabular} & \begin{tabular}[c]{@{}c@{}}CVACT-\textgreater \!\\ CVUSA\end{tabular} & \begin{tabular}[c]{@{}c@{}}CVUSA-\textgreater \!\\ CVACT\end{tabular} \\ \hline
unknown r@1          & 80.41\%                                                            & 85.39\%                                                            & 12.84\%                                                            & 19.43\%                                                            \\ \hhline{|=|=|=|=|=|}
mean error (all)     & 4.79$^\circ$                                                               & 3.65$^\circ$                                                               & 18.28$^\circ$                                                              & 13.00$^\circ$                                                              \\ 
mean error (matched)\! & 1.85$^\circ$                                                               & 1.96$^\circ$                                                               & 2.98$^\circ$                                                               & 2.27$^\circ$                                                               \\ \hhline{|=|=|=|=|=|}
r@2 (all)            & 72.60\%                                                            & 83.87\%                                                            & 57.09\%                                                            & 55.94\%                                                            \\ 
r@2 (matched)        & 77.03\%                                                            & 86.15\%                                                            & 70.20\%                                                            & 65.30\%                                                            \\
r@2 (matched to all)\! & 61.94\%                                                            & 73.56\%                                                            & 9.02\%                                                             & 12.69\%                                                            \\ \hline
\end{tabular}
\vspace{-2mm}
\end{table}

\begin{table}[]
\small
\caption{Performance of DSM~\cite{ShiEtAl2020}$^*$, DSM (all), and models with AL (angle loss), FI and CS.}
\label{tab:nofix vs angle vs FI CS}
\begin{tabular}{|l|cccc|}
\hline
Model               & \multicolumn{1}{c|}{\begin{tabular}[c]{@{}c@{}}known\\ r@1 (\%)\end{tabular}} & \multicolumn{1}{c|}{\begin{tabular}[c]{@{}c@{}}unknown\\ r@1 (\%)\end{tabular}} & \multicolumn{1}{c|}{\begin{tabular}[c]{@{}c@{}}Mean\\error$\downarrow$\end{tabular} }    & {\begin{tabular}[c]{@{}c@{}}r@2$^\circ$\\(\%)\end{tabular}}      \\ \hline
                    & \multicolumn{4}{c|}{CVUSA}                                                                                                                                                                                            \\ \hline
DSM~\cite{ShiEtAl2020}*                 & \multicolumn{1}{c|}{93.29}                                                    & \multicolumn{1}{c|}{80.54}                                                      & \multicolumn{1}{c|}{5.29$^\circ$}          & 47.49          \\ \hline
DSM (all)           & \multicolumn{1}{c|}{94.99}                                                    & \multicolumn{1}{c|}{84.30}                                                      & \multicolumn{1}{c|}{4.53$^\circ$}          & 50.76          \\ \hline
DSM (all) + AL      & \multicolumn{1}{c|}{94.90}                                                    & \multicolumn{1}{c|}{84.76}                                                      & \multicolumn{1}{c|}{4.76$^\circ$}          & 47.20          \\ \hline
DSM (all) + AL + FI & \multicolumn{1}{c|}{\textbf{95.29}}                                           & \multicolumn{1}{c|}{84.98}                                                      & \multicolumn{1}{c|}{3.78$^\circ$}          & 82.13          \\ \hline
DSM (all) + AL + CS & \multicolumn{1}{c|}{95.27}                                                    & \multicolumn{1}{c|}{\textbf{85.19}}                                             & \multicolumn{1}{c|}{\textbf{3.77$^\circ$}} & \textbf{82.42} \\ \hline
                    & \multicolumn{4}{c|}{CVACT}                                                                                                                                                                                            \\ \hline
DSM~\cite{ShiEtAl2020}*                 & \multicolumn{1}{c|}{83.90}                                                    & \multicolumn{1}{c|}{75.14}                                                      & \multicolumn{1}{c|}{6.26$^\circ$}          & 44.13          \\ \hline
DSM (all)           & \multicolumn{1}{c|}{86.06}                                                    & \multicolumn{1}{c|}{80.09}                                                      & \multicolumn{1}{c|}{5.67$^\circ$}          & 46.07          \\ \hline
DSM (all) + AL      & \multicolumn{1}{c|}{86.15}                                                    & \multicolumn{1}{c|}{80.04}                                                      & \multicolumn{1}{c|}{5.56$^\circ$}          & 45.25          \\ \hline
DSM (all) + AL + FI & \multicolumn{1}{c|}{\textbf{86.59}}                                           & \multicolumn{1}{c|}{\textbf{80.38}}                                             & \multicolumn{1}{c|}{4.88$^\circ$}          & 70.87          \\ \hline
DSM (all) + AL + CS & \multicolumn{1}{c|}{86.29}                                                    & \multicolumn{1}{c|}{80.23}                                                      & \multicolumn{1}{c|}{\textbf{4.75$^\circ$}} & \textbf{72.28} \\ \hline
\end{tabular}
\end{table}

\subsubsection{Limited FOV}

\begin{table}[]
\small
\tabcolsep=2.12mm
\vspace{-2mm}
\caption{Performance of models trained on 360$^\circ$ CVACT images and tested on 180$^\circ$, 90$^\circ$. The absolute improvements to DSM~\cite{ShiEtAl2020}$^*$ are given in the brackets.}
\label{tab:360 to limit}
\vspace{-2mm}
\begin{tabular}{|c|c|c|c|c|}
\hline
FOV & r@1 (\%)     & Mean error $\downarrow$  & r@2$^\circ$ (\%)      & r@5$^\circ$ (\%)     \\ \hline
180$^\circ$ & 56.47 (9.60)  & 9.48$^\circ$ (2.79$^\circ$)  & 51.12 (19.56) & 81.85 (8.91) \\ \hline
90$^\circ$  & 21.88 (6.83) & 21.57$^\circ$ (3.62$^\circ$) & 30.90 (10.92)  & 55.35 (6.64) \\ \hline
\end{tabular}
\vspace{-4mm}
\end{table}

We test on images with 180$^\circ$ (fish-eye camera) and 90$^\circ$ (wide-angle camera). The first test uses models trained on 360$^\circ$ images to simulate the situation when a model is trained on well collected 360$^\circ$ images, but the test images are crowdsourced with limited FOV. \autoref{tab:360 to limit} shows the average performances of FI models on CVACT. The absolute improvements to our re-implemented baseline DSM~\cite{ShiEtAl2020}$^*$ are given in the brackets. For geo-localization unknown r@1, our result has a 9.60\% and 6.83\% improvement for 180$^\circ$ and 90$^\circ$ tests. Recent work SEH-VGG16 ~\cite{GuoEtAl2022a} reports r@1 of 45.99\% and 14.06\%  on the same test. For the orientation estimation, we observe improvement on all metrics compared to DSM~\cite{ShiEtAl2020}$^*$. However, the performance and the absolute improvements on most of metrics gradually reduced when the FOV is more restricted. The mean error for 90$^\circ$ FOV is nearly one fourth of its FOV and the r@2$^\circ$ declines to less than 50\%. This observation is aligned with our intuition that the difficulty of precise geo-localization and orientation estimation increases drastically when the FOV decreases.

Secondly, we test on FI models trained with limited FOV to understand whether images with limited FOV still contain enough information for learning fine-grained orientation estimation. \autoref{tab:limited FOV trained } present the results of models trained on 180$^\circ$ and 90$^\circ$. Comparing to the baseline, our model FI gives improvement in most of the metrics, indicating our methods is applicable for limited FOV. The improvement to baselines and performance itself on 180$^\circ$ is much more significant than 90$^\circ$. This reduction caused by decreased FOV is more drastic in models trained with limited FOV compared to what we observed in full FOV trained models (\autoref{tab:360 to limit}). Additionally, besides the unknown r@1 of 180$^\circ$ obtains a higher performance than the model trained with 360$^\circ$, all other metrics in limited FOV training result in worse performance, especially on orientation extraction. It indicates that missing partial FOV has great influence on both orientation and location extraction. However, its impacts on finding precise image orientation might be more significant than location extraction.

\begin{table}[]

\small
\caption{CVACT model trained and tested on limited FOV.}
\label{tab:limited FOV trained }
\vspace{-0.8em}
\begin{tabular}{|c|c|c|c|c|c|}
\hline
FOV                  & Model     & r@1 (\%) & Mean error $\downarrow$ & r@2$^\circ$ (\%) & r@5$^\circ$ (\%) \\ \hline
\multirow{2}{*}{180$^\circ$} & DSM~\cite{ShiEtAl2020}$^*$       & 58.11\%  & 12.31$^\circ$      & 25.41\%  & 64.13\%  \\ \cline{2-6} 
                     & Ours FI & 67.07\%  & 10.26$^\circ$      & 34.82\%  & 69.61\%  \\ \hline
\multirow{2}{*}{90$^\circ$}  & DSM~\cite{ShiEtAl2020}$^*$       & 16.77\%  & 39.50$^\circ$      & 8.22\%   & 23.61\%  \\ \cline{2-6} 
                     &  Ours FI & 21.08\%  & 40.42$^\circ$      & 9.01\%   & 21.23\%  \\ \hline
\end{tabular}
\vspace{-1em}
\end{table}

\section{Findings and Discussion}
(1) Orientation estimation: our methods improve the orientation estimation performance for both datasets. It has a higher influence on CVUSA, as images in CVUSA have less complex scenes and obvious features. In general, CS obtains slightly better results in our test cases, however, FI gives more fundamental fine-grained orientation curve generation, this could be useful when prior knowledge of the rough orientation is available. (2) Geo-localization: by integrating fine-grained orientation estimation, the trained models obtain higher performance, compared to baseline models. The r@1 for both datasets achieve better scores than existing methods. Note that our initiatives on integrating fine-grained orientation estimation are compatible with some recent works, \eg \cite{GuoEtAl2022a} on sampling strategies. (3) Evaluation: When the street-view images are correctly geo-localized, the orientation estimation has higher precision. However, most of the location-incorrect images still obtain high to medium quality orientation estimation. It is also fairer to evaluate orientation estimation independently from geo-localization. (4) Limited FOV: When a model is trained on 360$^\circ$ images and is tested on limited FOV, the performance of geo-localization and orientation estimation both decrease. The more limited the FOV restricts, the worse the performance is. When a model is trained and tested on a limited FOV, the missing information has higher impacts on fine-grained orientation estimation than geo-localization. 

\section{Related Work}

Because of its unique prospective and large ground coverage, satellite imagery has been used in different geo-information extraction applications~\cite{yin2021multimodal, yin2022context, ZhouEtAl2018, DemirEtAl2018, BastaniEtAl2018,BastaniEtAl2018b, hu2021geopalette}. In recent years, satellite imagery-based cross-view matching has become one of the mainstream approaches for street-view imagery geo-localization~\cite{WorkmanEtAl2015, VoHays2017, HuEtAl2018, CaiEtAl2019, LiuLi2019, RegmiShah2019, ShiEtAl2019, ShiEtAl2019a, SunEtAl2019a, RodriguesTani2020,ShiEtAl2020, VerdeEtAl2020, TokerEtAl2021, WangEtAl2021, YangEtAl2021c, ZhuEtAl2020, ZhangEtAl2022, GuoEtAl2022a, ZhuEtAl2022}. It partially solves the task to prepare crowdsourced street-view images for ``ready-to-use'' status. The main challenge of image-based cross-view matching is to bridge the gap between the two views. Traditional key-point extraction methods, \eg SIFT~\cite{Lowe2004}, do not work on images taken from drastically different viewing angles \cite{DeToneEtAl2018,TianEtAl2017,AltwaijryEtAl2016}. Vo \etal~\cite{VoHays2017} propose a deep learning based method to extract features from different views by studying triplets. Follow-up works aim to further reduce the gaps between the two views and their contributions can be categorized into three aspects. 1) Enhancing correspondences between two views by introducing new learnable components in the network or transforming images from one view to another view~\cite{HuEtAl2018, CaoEtAl2019, LiuLi2019, WangEtAl2021, ShiEtAl2019, ShiEtAl2019a, YangEtAl2021c, ZhangEtAl2022, ShiEtAl2019a, RegmiShah2019, TokerEtAl2021, ZhuEtAl2022, ShiEtAl2022}. Advanced techniques, such as NetVlAD \cite{arandjelovic2016netvlad}, spatial transformer layers~\cite{JaderbergEtAl2016}, generative adversarial models~\cite{GoodfellowEtAl2014}, attention or vision transformer mechanics \cite{VaswaniEtAl2017, DosovitskiyEtAl2020} are tested. A major breakthrough realized by polar transformation is introduced in \cite{ShiEtAl2019a}. 2) Using data sampler/additional information/augmentations to guide the training \cite{HuEtAl2018,CaiEtAl2019,GuoEtAl2022a, LiuLi2019, RodriguesTani2020}. 3) Utilizing orientation estimation. \cite{VoHays2017} adds an auxiliary loss for angle orientation. \cite{ShiEtAl2020, ZhuEtAl2020, ZhaiEtAl2017} find the orientation by correlating extracted features from two views.

Researchers are also looking for more challenging problems. The University-1652 dataset \cite{ZhengEtAl2020} has been introduced to include UAV data as one of the platforms for geo-localization. The VIGOR dataset \cite{ZhuEtAl2021} is provided to refine the precision of geo-localization beyond image retrieval for large scale searching. Additionally, cross-view matching is also adapted for small scale search in recent works, such as in autonomous driving domain \cite{XiaEtAl2021, ShiLi2022} to refine the location and/or orientation when prior knowledge is given.

\section{Conclusion and future work}
In this work, we formally define the fine-grained orientation estimation of street-view images by cross-view matching with satellite imagery. The aim is to extend the scope of current work on geo-localization and give it a more realistic setting and meaningful goal to fully explore the potential of crowdsourced street-view images. We achieve 82.4\% and 72.3\% accuracy for images with orientation estimation error below 2$^\circ$ for CVUSA and CVACT. Our results also show that by integrating fine-grained orientation estimation, the performance of geo-localization are improved. 

Concurrent works also pursue new initiatives in different directions. We highlight the following potential aspects for future works: (1) Limited FOV. In this work, we have tested geo-localization and orientation estimation with different settings. Among all, finding geo-localization and fine-grained orientation from images with limited FOV is the most challenging problem but also the one with the highest societal impacts.
(2) Combining with other data sources. Map rendered imagery can be an additional input or a replacement for satellite imagery. Compared to satellite imagery, map rendered images contain fewer details, but are easier to obtain and have less influence on seasonal/weather conditions. When prior knowledge of location/orientation is available, additional data sources, \eg GPS, can be used to reduce the searching space for local scale search. (3) Use of metrics. In this work, we evaluate the performance of all test images. However, future works can further explore other extensions of the existing metrics to cater to different task-specific requirements. (4) Beyond image retrieval. Our work assumes satellite images can be densely cropped in the area of interest, hence the camera locations are not far away from the satellite image centers. However, fine-grained location extraction beyond image retrieval is definitely helpful to reduce the sampling density of the satellite images for large scale applications. The new challenge is to extract fine-grained location and orientation at the same time for off-center orientation unknown images without any prior knowledge.

\begin{acks}
This work was funded by the Grab-NUS AI Lab, a joint collaboration between GrabTaxi Holdings Pte. Ltd. and National University of Singapore, and the Industrial Postgraduate Program (Grant: S18-1198-IPP-II) funded by the Economic Development Board of Singapore.
\end{acks}

\clearpage
\bibliographystyle{ACM-Reference-Format}
\bibliography{main}

\clearpage


\appendix

\section{Supplementary Materials}
\subsection{Method to rotate and crop street-view image} \label{alg:shift and rotate}

The \autoref{alg:shift} shows the augmentation to rotate the street-view images and create orientation shift ground truth in feature space. It is adapted from DSM~\cite{ShiEtAl2020}, while allowing sub-pixel locations of the ground truth for fine-grained orientation estimation.
\begin{algorithm}[!h]
\caption{Shift and crop Street-view Image}
\label{alg:shift}
\small
\KwIn{Street-view image $I$, feature map width $w$, field of view $\gamma$.}
\KwOut{Shifted image $I_{sc}$, ground truth alignment $w_\text{gt}$.}

$w_f = \gamma / 360 * w$   //calculate the cropped image width

$x_\text{shift} = \text{randint}(0, I.\text{width}-1)$ //define a random shift

\tcc{shift image}
$I_s = \text{concat}(I[I.\text{width}-x_\text{shift}:], I[:I.\text{width}-x_\text{shift}])$

$I_{sc} = I_s[:w_f]$ // cropped the first $w_f$ pixels

$w_\text{gt} = \text{mod}(((I.\text{width} - x_\text{shift})/ I.\text{width} * w), w)$ //shift in feature map width

\Return{ $I_{sc}$, $w_\text{gt}$}
\end{algorithm}

\subsection{Polar transformation}
 We used the polar transformation from DSM~\cite{ShiEtAl2020} to pre-process the satellite images. Given a satellite image with the size of $(S_s, S_s)$, to transform it into a flattened image with the size as street-view image $(H_g, W_g)$, the pixel relation between satellite coordinate ${x_i^s, y_i^s}$ and street-view coordinate ${x_i^g, y_i^g}$ (both coordinate take the left-upper corner as the origin): 
    \begin{displaymath} \label{eq:polar}
        \small
        \begin{split}
            x_i^s = \frac{S_s}{2} - \frac{S_s}{2} \frac{H_g-y_i^g}{H_g}\sin(\frac{2\pi}{W_g}x_i^g) \\
            y_i^s = \frac{S_s}{2} + \frac{S_s}{2} \frac{H_g-y_i^g}{H_g}\cos(\frac{2\pi}{W_g}x_i^g)
        \end{split}
    \end{displaymath}

\begin{figure}[!b]
  \includegraphics[width=0.47 \textwidth, trim = 0cm 16cm 42cm 0cm, clip]{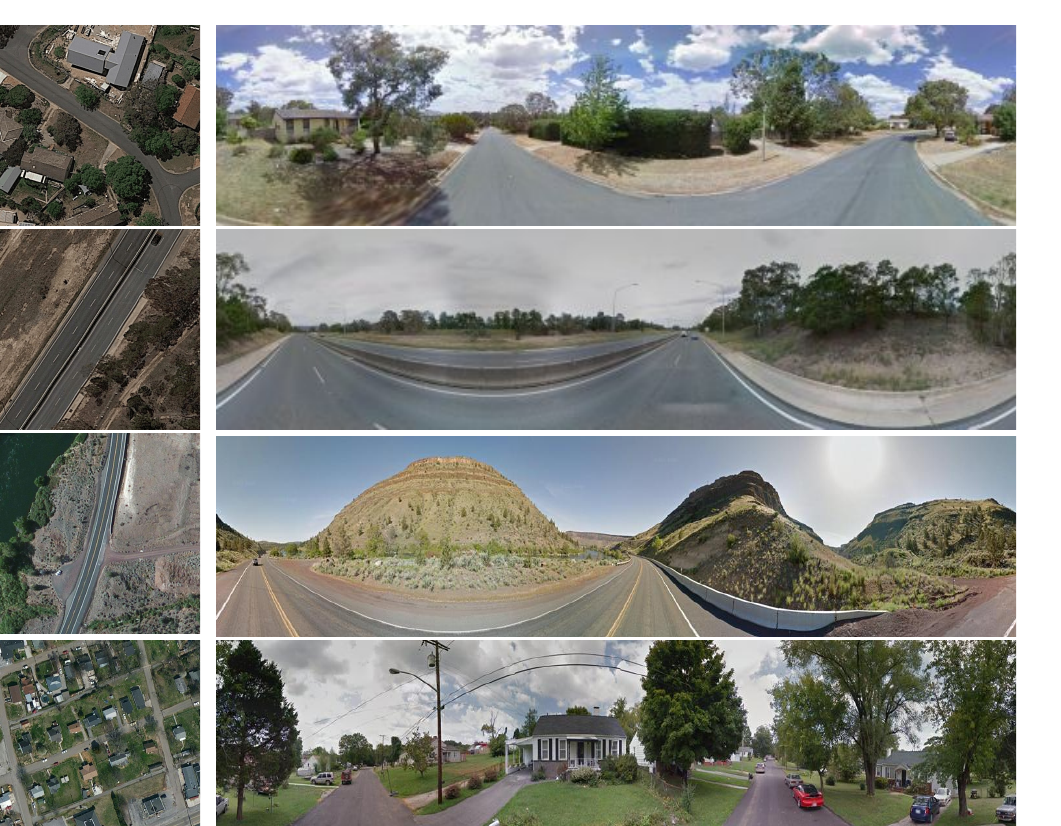}
  \caption{Sample images from CVUSA~\cite{WorkmanEtAl2015} and CVACT~\cite{LiuLi2019} dataset. Top 2: CVACT; Bottom 2: CVUSA.}
  \label{fig:samples}
\end{figure}
\subsection{Sample images for CVACT and CVUSA}
Sample images from CVACT/CVUSA are shown in \autoref{fig:samples}. Note that the two datasets are collected in the US and Australia separately and have a non-negligible domain shift. CVUSA contains a mix of commercial, residential, suburban, and rural areas and CVACT leans towards urban/suburban styles. Moreover, the satellite images in CVUSA have a higher ground coverage but a lower resolution than CVACT, introducing another substantial domain shift.

\vspace{-0.5em}
\subsection{Histograms for same/across dataset tests} \label{section:additional hist}
\begin{figure}[!h]
  \vspace{-1em}
  \includegraphics[width=0.47 \textwidth, trim = 0.4cm 20.3cm 42cm 0cm, clip]{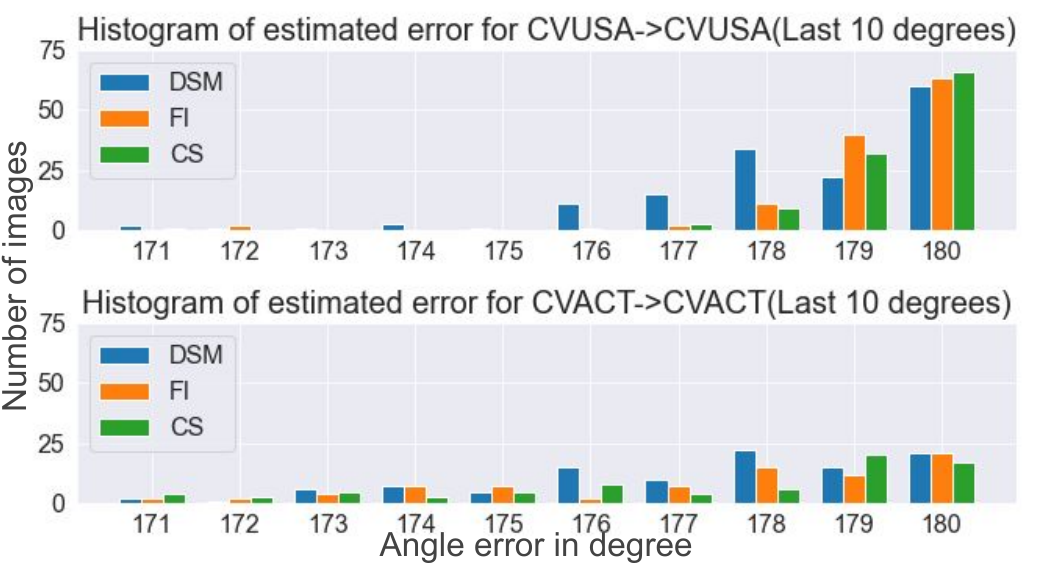}
  \vspace{-1.3em}
  \caption{Histogram of angle errors for CVUSA and CVACT on same dataset tests (last 10 degrees).}
  \label{fig:last10}\vspace{-2em}
\end{figure}

\begin{figure}[!h]
  \includegraphics[width=0.47 \textwidth, trim = 0.4cm 20.3cm 42cm 0cm, clip]{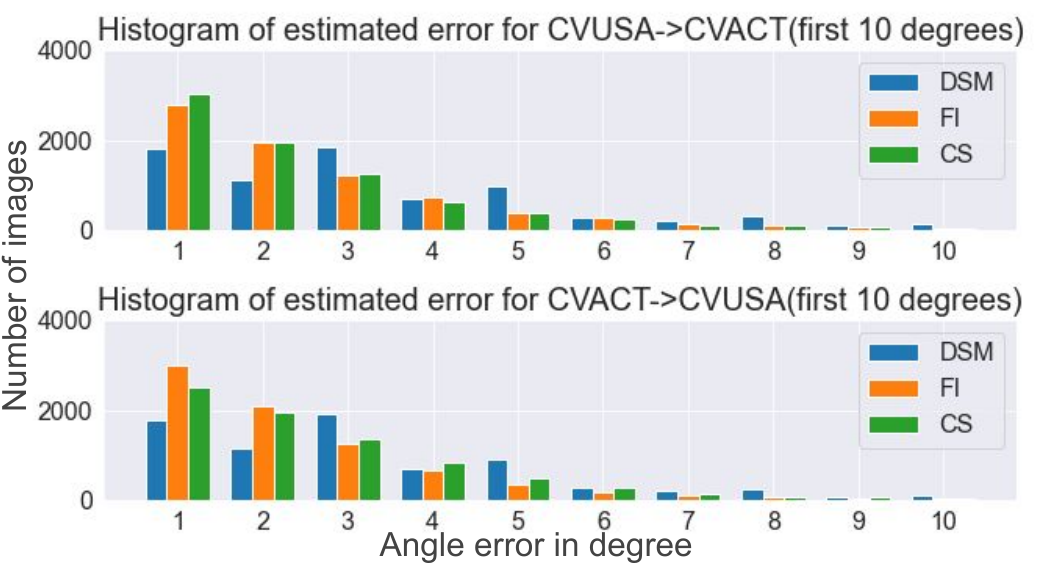}
  \vspace{-1.3em}
  \caption{Histogram of angle errors for CVUSA and CVACT on across dataset tests (first 10 degrees).}
  \label{fig:cross_hist}\vspace{-2em}
\end{figure}

\begin{figure}[!h]
  \includegraphics[width=0.47 \textwidth, trim = 0.4cm 20.3cm 42cm 0cm, clip]{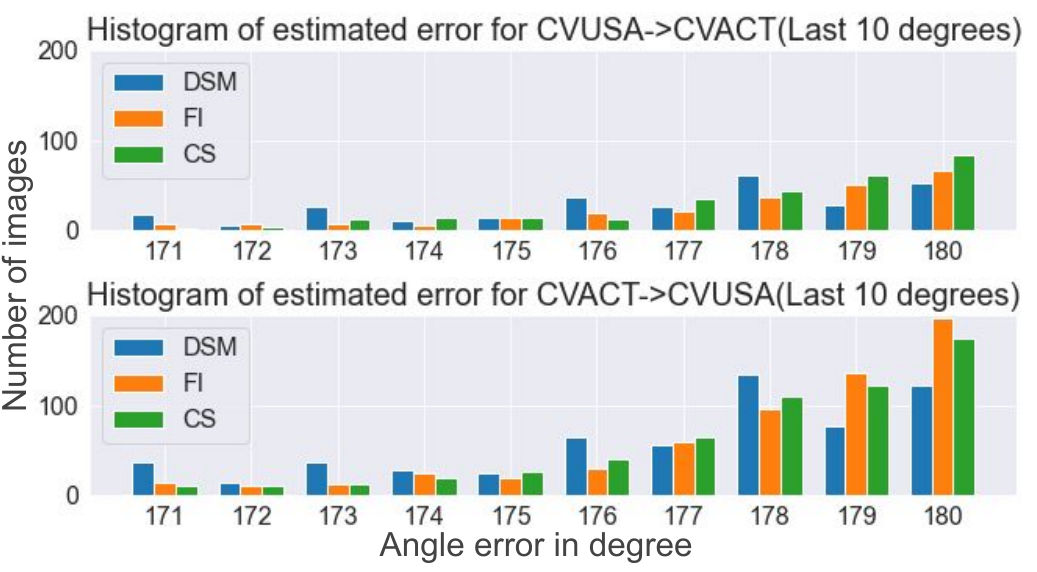}
  \vspace{-1.3em}
  \caption{Histogram of angle errors for CVUSA and CVACT on across dataset tests (last 10 degrees).}
  \label{fig:cross_last10}
\end{figure}

\autoref{fig:last10}, \autoref{fig:cross_hist} and \autoref{fig:cross_last10} shows the histograms of the best models of our re-implemented DSM~\cite{ShiEtAl2020}$^*$, FI and CS for same dataset test (last 10$^\circ$) and across dataset test (first and last 10$^\circ$).

\subsection{Performance on across dataset tests} \label{section:across}
For across dataset tests, we train on one dataset and test on the other dataset. For example, in \autoref{tab: across test angle} and \autoref{tab: across test location} CVACT $\rightarrow$ CVUSA means model trained on CVACT train set and tested on CVUSA test set.

The performance on orientation extraction is shown in \autoref{tab: across test angle}. Because of the domain shift between the two datasets, both test cases have larger mean errors to begin with, compared to same dataset tests. The r@2$^\circ$ have an absolute improvement around 20\% to 23\% from the original accuracy about 32\% for both cases. CVACT $\rightarrow$ CVUSA obtained better improvements on the mean error (about 3.6$^\circ$) and r@5$^\circ$ (about 9.7\%). Combining the fact that CVACT has a worse initial mean error for across dataset tests, we believe the higher ratio of urban images in CVACT leads the models to leverage the obvious features, \eg buildings. When such features are missing in the CVUSA, it results in a worse performance than the reversed case (CVUSA$\rightarrow$CVACT). By integrating fine-grained orientation estimation in training, the models improve on generalization and transferability and have less dependency on easy features.

For geo-localization, by learning fine-grained orientation estimation, the trained models gain higher generalization and transferability in across dataset tests without implementing additional sampling strategy or computational expensive architecture. Compared to our implementation of the baseline model DSM~\cite{ShiEtAl2020}$^*$, the r@1 for known/unknown orientation for CVUSA$\rightarrow$CVACT are improved by 5.11\%, 2.43\% for FI and 3.57\%, 1.01\% for CS; for CVACT$\rightarrow$CVUSA are improved by 8.61\%,	4.83\% for FI and 7.38\%, 4.46\% for CS. Compared to recent works, L2LTR\cite{YangEtAl2021c} achieves higher performance on known orientation tests, with the usage of ResNet backbone with 12 layers of vision transformers on each view to bridge the gap between two views.

\begin{table}[]
\small
\caption{Orientation extraction results in across dataset tests.}

\label{tab: across test angle}
\begin{tabular}{|c|ccc|ccc|}
\hline
Model                                                    & \multicolumn{1}{c|}{\begin{tabular}[c]{@{}c@{}}Mean\\ error $\downarrow$\end{tabular}}            & \multicolumn{1}{c|}{\begin{tabular}[c]{@{}c@{}}r@2$^\circ$ \\ (\%)\end{tabular}}               & \begin{tabular}[c]{@{}c@{}}r@5$^\circ$ \\ (\%)\end{tabular}              & \multicolumn{1}{c|}{\begin{tabular}[c]{@{}c@{}}Mean\\ error\end{tabular}}            & \multicolumn{1}{c|}{\begin{tabular}[c]{@{}c@{}}r@2$^\circ$\\ (\%)\end{tabular}}               & \begin{tabular}[c]{@{}c@{}}r@5$^\circ$\\ (\%)\end{tabular}              \\ \hline
                                                         & \multicolumn{3}{c|}{CVUSA-\textgreater{}CVACT}                                                                                                                                                                                                 & \multicolumn{3}{c|}{CVACT-\textgreater{}CVUSA}                                                                                                                                                                                                 \\ \hline
DSM~\cite{ShiEtAl2020}*                                                      & \multicolumn{1}{c|}{15.48$^\circ$}                                                           & \multicolumn{1}{c|}{32.45}                                                            & 72.18                                                           & \multicolumn{1}{c|}{22.21$^\circ$}                                                           & \multicolumn{1}{c|}{32.40}                                                            & 72.04                                                           \\ \hline
\begin{tabular}[c]{@{}c@{}}Ours (FI)\\ Gain\end{tabular} & \multicolumn{1}{c|}{\begin{tabular}[c]{@{}c@{}}13.15$^\circ$\\ (2.33$^\circ$)\end{tabular}}          & \multicolumn{1}{c|}{\begin{tabular}[c]{@{}c@{}}52.94\\ (20.49)\end{tabular}}          & \begin{tabular}[c]{@{}c@{}}79.77\\ (7.59)\end{tabular}          & \multicolumn{1}{c|}{\begin{tabular}[c]{@{}c@{}}19.13$^\circ$\\ (3.08$^\circ$)\end{tabular}} & \multicolumn{1}{c|}{\textbf{\begin{tabular}[c]{@{}c@{}}54.95\\ (22.55)\end{tabular}}} & \begin{tabular}[c]{@{}c@{}}81.21\\ (9.17)\end{tabular}          \\ \hline
\begin{tabular}[c]{@{}c@{}}Ours (CS)\\ Gain\end{tabular} & \multicolumn{1}{c|}{\textbf{\begin{tabular}[c]{@{}c@{}}13.02$^\circ$\\ (2.46$^\circ$)\end{tabular}}} & \multicolumn{1}{c|}{\textbf{\begin{tabular}[c]{@{}c@{}}54.30\\ (21.85)\end{tabular}}} & \textbf{\begin{tabular}[c]{@{}c@{}}80.21\\ (8.03)\end{tabular}} & \multicolumn{1}{c|}{\textbf{\begin{tabular}[c]{@{}c@{}}18.61$^\circ$\\ (3.60$^\circ$)\end{tabular}}}          & \multicolumn{1}{c|}{\begin{tabular}[c]{@{}c@{}}54.13\\ (21.73)\end{tabular}}          & \textbf{\begin{tabular}[c]{@{}c@{}}81.73\\ (9.69)\end{tabular}} \\ \hline
\end{tabular}
\end{table}

\begin{table}[]
\caption{Geo-localization results in across dataset tests. *SAFA\cite{ShiEtAl2019a} across dataset result from \cite{YangEtAl2021c}.}

\label{tab: across test location}
\begin{tabular}{|c|cc|cc|}
\hline
Model     & \multicolumn{1}{c|}{\begin{tabular}[c]{@{}c@{}}known\\ r@1 (\%)\end{tabular}} & \begin{tabular}[c]{@{}c@{}}unknown\\ r@1 (\%)\end{tabular} & \multicolumn{1}{c|}{\begin{tabular}[c]{@{}c@{}}known\\ r@1 (\%)\end{tabular}} & \begin{tabular}[c]{@{}c@{}}unknown\\ r@1 (\%)\end{tabular} \\ \hline
      & \multicolumn{2}{c|}{CVUSA-\textgreater{}CVACT}                                                                                             & \multicolumn{2}{c|}{CVACT-\textgreater{}CVUSA}                                                                                             \\ \hline
SAFA~\cite{ShiEtAl2019a}      & \multicolumn{1}{c|}{30.40}                                                    & -                                                          & \multicolumn{1}{c|}{21.45}                                                    & -                                                          \\ \hline
DSM~\cite{ShiEtAl2020}*          & \multicolumn{1}{c|}{38.47}                                                    & 18.42                                                      & \multicolumn{1}{c|}{21.68}                                                    & 8.01                                                       \\ \hline
L2LTR~\cite{YangEtAl2021c}      & \multicolumn{1}{c|}{\textbf{47.55}}                                           & -                                                          & \multicolumn{1}{c|}{\textbf{33.00}}                                           & -                                                          \\ \hhline{|=|=|=|=|=|}
Ours (FI) & \multicolumn{1}{c|}{43.58}                                                    & \textbf{20.85}                                             & \multicolumn{1}{c|}{30.29}                                                    & \textbf{12.84}                                             \\ \hline
Ours (CS) & \multicolumn{1}{c|}{42.04}                                                    & 19.43                                                      & \multicolumn{1}{c|}{29.06}                                                    & 12.47                                                      \\ \hline
\end{tabular}
\end{table}
\subsection{Performance on mixed dataset tests} \label{section:mixed result}
\autoref{tab: mixed test angle} and \autoref{tab: mixed test location} show the performance of best instance of each model trained on CVACT in a test that the test set of the two datasets are mixed. Besides the metrics used in the main sessions, few new metrics are added to understand the improvements in the sets and their contribution to the overall performance. For r@1, r@2$^\circ$, r@5$^\circ$ and mean error, we show the breakdown for same dataset (data in CVACT test set) and across dataset (data in CVUSA test set). Additionally, the a hit rate within its own dataset is calculated, which shows the percentage of queries of each dataset choose a satellite candidate from its own satellite pool. 

Compared to DSM~\cite{ShiEtAl2020}*: (1) our FI model not only improves the performance of the same test dataset, but also propagates the improvement to across test dataset. For r@5$^\circ$, mean error, known r@1 and all hit rates, the gains in the across test set are even higher than the same test set. (2) The hit rate for unknown and known orientation tests both obtain more balanced performances and reduce the gap between the same and across test datasets. Our model gives less unfair favors to the training dataset. Both of the results show that learning orientation extraction improves not only the performance of the trained models, but also improves the generalization of the models to be applied in other geo-location/slightly different acquisition settings.

\begin{table}[!h]
\caption{Orientation extraction performance of mixed dataset tests on model trained on CVACT.}
\label{tab: mixed test angle}
\begin{tabular}{|l|c|c|c|}
\hline
                    & Ours FI & DSM~\cite{ShiEtAl2020}* & Gain    \\ \hline
r@2$^\circ$                 & 64.85\%  & 38.70\%   & 26.15\% \\ 
r@2$^\circ$ (same)          & 72.60\%  & 44.26\%   & 28.34\% \\ 
r@2$^\circ$ (across)        & 57.10\%  & 33.15\%   & 23.95\% \\ \hline
r@5$^\circ$                 & 87.45\%  & 80.60\%   & 6.85\%  \\ 
r@5$^\circ$ (same)          & 92.54\%  & 88.44\%   & 4.10\%  \\ 
r@5$^\circ$ (across)        & 82.36\%  & 72.76\%   & 9.60\%  \\ \hline
Mean error $\downarrow$          & 11.53$^\circ$    & 14.22$^\circ$       & 2.69$^\circ$      \\ 
Mean error (same) $\downarrow$  & 4.79$^\circ$       & 6.30$^\circ$        & 1.51$^\circ$      \\ 
Mean error (across) $\downarrow$ & 18.28$^\circ$      & 22.14$^\circ$       & 3.86$^\circ$      \\ \hline
\end{tabular}
\end{table}

\begin{table}[!h]
\caption{Geo-localization performance of mixed dataset tests on model trained on CVACT.}
\label{tab: mixed test location}
\begin{tabular}{|l|c|c|c|}
\hline
\textbf{}            & Ours FI & DSM~\cite{ShiEtAl2020}* & Gain   \\ \hline
unknown r@1          & 46.21\%  & 40.99\%   & 5.22\% \\ 
unknown r@1 (same)   & 79.90\%  & 74.56\%   & 5.34\% \\ 
unknown r@1 (across) & 12.53\%  & 7.42\%    & 5.11\% \\ \hline
unknown hit (same)   & 97.57\%  & 97.13\%   & 0.44\% \\ 
unknown hit (across) & 95.42\%  & 88.65\%   & 6.76\% \\ \hline
known r@1            & 58.05\%  & 51.96\%   & 6.09\% \\ 
known r@1 (same)     & 86.37\%  & 83.35\%   & 3.02\% \\ 
known r@1 (across)   & 29.73\%  & 20.57\%   & 9.16\% \\ \hline
known hit (same)     & 98.20\%  & 97.87\%   & 0.33\% \\ 
known hit (across)   & 96.69\%  & 91.69\%   & 5.00\% \\ \hline
\end{tabular}
\end{table}

\subsection{Large location offset test} \label{section:vigor result}
In this work, we consider the matched pairs in CVUSA and CVACT are perfectly location aligned. However, the datasets do contain small location translation offsets due to GPS errors \eg last row of \autoref{fig:cvact fi worst}, the camera location shall be on the main road instead of the side road, the top 1 prediction is actually closer to the real position. For some extreme cases, the camera locations of the street-view images are off their matched satellite images. e.g. first, second and fourth row of \autoref{fig:cvact fi worst}, the actual camera location drifts off from the matched image (ground truth) and actually is around the top 1 prediction given by our model. Hence, our models tolerate small location offsets.

Although it is out of the scope of this work, we test on the VIGOR (camera locations have large offsets to the satellite image centers) out of curiosity. The result is shown in \autoref{tab: vigor test}. Large offsets indeed demolish the overall performance. But, we found adding fine-grained orientation extraction still improves the performance:
\begin{itemize}[leftmargin=*]
    \item Training with unknown orientation increases performance on r@1, r@2$^\circ$, r@5$^\circ$ and mean orientation error, compared to models trained with known orientation.
    \item Training with a high scaling factor of fine-grained orientation improves the fine-grained orientation extraction, compared to models trained with a lower scaling factor.
\end{itemize}

\begin{table}[]
\caption{Model performance on VIGOR dataset with our CS methods.}
\label{tab: vigor test}
\begin{tabular}{|c|c|c|c|}
\hline
                 & S10 known & S1 unknown & S10 unknown \\ \hline
known r@1        & 16.99\%    & 20.81\%     & 20.81\%      \\ \hline
known hit rate   & 22.86\%    & 28.60\%     & 28.56\%      \\ \hline
unknown r@1      & 11.47\%    & 16.67\%     & 16.70\%      \\ \hline
unknown hit rate & 16.34\%    & 24.54\%     & 24.46\%      \\ \hline
Mean error $\downarrow$         & 36.46$^\circ$       & 34.53$^\circ$        & 34.51$^\circ$         \\ \hline
r@2$^\circ$              & 10.54\%    & 9.80\%      & 11.52\%      \\ \hline
r@5$^\circ$               & 24.21\%    & 27.23\%     & 26.43\%      \\ \hline
\end{tabular}
\end{table}

\clearpage
\onecolumn
\subsection{Performance of geo-localization and orientation on two datasets with different configurations.} \label{session: full compare}
\autoref{tab: full performance} shows the performance with different configurations. The average performance of three instances is reported. In the main paper, angle weight of 0.3 is used.

\begin{table*}[h!]
\caption{Performance of geo-localization and orientation on two datasets with different configurations. }
\label{tab: full performance}
\begin{tabular}{|c|c|ccccc|ccccc|}
\hline
                    &                                                        & \multicolumn{5}{c|}{CVACT -\textgreater CVACT}                                                                                                                                                                                                                             & \multicolumn{5}{c|}{CVACT-\textgreater CVUSA}                                                                                                                                                                                                                              \\ \hline
                    & \begin{tabular}[c]{@{}c@{}}angle\\ weight\end{tabular} & \multicolumn{1}{c|}{\begin{tabular}[c]{@{}c@{}}known\\ r@1\end{tabular}} & \multicolumn{1}{c|}{\begin{tabular}[c]{@{}c@{}}unknown\\ r@1\end{tabular}} & \multicolumn{1}{c|}{\begin{tabular}[c]{@{}c@{}}Mean\\ error ($^\circ$)$\downarrow$\end{tabular}} & \multicolumn{1}{c|}{r@2$^\circ$}    & r@5$^\circ$    & \multicolumn{1}{c|}{\begin{tabular}[c]{@{}c@{}}known\\ r@1\end{tabular}} & \multicolumn{1}{c|}{\begin{tabular}[c]{@{}c@{}}unknown\\ r@1\end{tabular}} & \multicolumn{1}{c|}{\begin{tabular}[c]{@{}c@{}}Mean\\ error($^\circ$)$\downarrow$\end{tabular}} & \multicolumn{1}{c|}{r@2$^\circ$}    & r@5$^\circ$    \\ \hline
DSM~\cite{ShiEtAl2020}*                 & -                                                      & \multicolumn{1}{c|}{83.90\%}                                             & \multicolumn{1}{c|}{75.14\%}                                               & \multicolumn{1}{c|}{6.26}                                                 & \multicolumn{1}{c|}{44.13\%} & 88.31\% & \multicolumn{1}{c|}{21.69\%}                                             & \multicolumn{1}{c|}{7.90\%}                                                & \multicolumn{1}{c|}{22.21}                                                & \multicolumn{1}{c|}{32.40\%} & 72.04\% \\ \hline
DSM (all)           & -                                                      & \multicolumn{1}{c|}{86.06\%}                                             & \multicolumn{1}{c|}{80.09\%}                                               & \multicolumn{1}{c|}{5.67}                                                 & \multicolumn{1}{c|}{46.07\%} & 89.91\% & \multicolumn{1}{c|}{25.79\%}                                             & \multicolumn{1}{c|}{11.00\%}                                               & \multicolumn{1}{c|}{20.96}                                                & \multicolumn{1}{c|}{35.05\%} & 75.85\% \\ \hline
DSM (all) + AL      & 0.1                                                    & \multicolumn{1}{c|}{86.12\%}                                             & \multicolumn{1}{c|}{80.19\%}                                               & \multicolumn{1}{c|}{5.73}                                                 & \multicolumn{1}{c|}{43.89\%} & 88.66\% & \multicolumn{1}{c|}{26.46\%}                                             & \multicolumn{1}{c|}{11.28\%}                                               & \multicolumn{1}{c|}{20.19}                                                & \multicolumn{1}{c|}{35.22\%} & 76.28\% \\ \hline
DSM (all) + AL      & 0.3                                                    & \multicolumn{1}{c|}{86.15\%}                                             & \multicolumn{1}{c|}{80.04\%}                                               & \multicolumn{1}{c|}{5.56}                                                 & \multicolumn{1}{c|}{45.25\%} & 89.42\% & \multicolumn{1}{c|}{26.19\%}                                             & \multicolumn{1}{c|}{11.08\%}                                               & \multicolumn{1}{c|}{20.57}                                                & \multicolumn{1}{c|}{35.12\%} & 76.08\% \\ \hline
DSM (all) + AL      & 0.5                                                    & \multicolumn{1}{c|}{86.17\%}                                             & \multicolumn{1}{c|}{80.14\%}                                               & \multicolumn{1}{c|}{5.67}                                                 & \multicolumn{1}{c|}{44.08\%} & 88.86\% & \multicolumn{1}{c|}{25.52\%}                                             & \multicolumn{1}{c|}{11.00\%}                                               & \multicolumn{1}{c|}{20.86}                                                & \multicolumn{1}{c|}{32.76\%} & 73.51\% \\ \hline
DSM (all) +FI       & -                                                      & \multicolumn{1}{c|}{86.50\%}                                             & \multicolumn{1}{c|}{80.53\%}                                               & \multicolumn{1}{c|}{4.82}                                                 & \multicolumn{1}{c|}{70.52\%} & 92.24\% & \multicolumn{1}{c|}{27.27\%}                                             & \multicolumn{1}{c|}{11.23\%}                                               & \multicolumn{1}{c|}{19.56}                                                & \multicolumn{1}{c|}{52.67\%} & 80.10\% \\ \hline
DSM (all) + AL + FI & 0.1                                                    & \multicolumn{1}{c|}{86.30\%}                                             & \multicolumn{1}{c|}{80.45\%}                                               & \multicolumn{1}{c|}{4.89}                                                 & \multicolumn{1}{c|}{69.92\%} & 92.20\% & \multicolumn{1}{c|}{27.79\%}                                             & \multicolumn{1}{c|}{11.41\%}                                               & \multicolumn{1}{c|}{18.56}                                                & \multicolumn{1}{c|}{53.65\%} & 81.02\% \\ \hline
DSM (all) + AL + FI & 0.3                                                    & \multicolumn{1}{c|}{86.59\%}                                             & \multicolumn{1}{c|}{80.38\%}                                               & \multicolumn{1}{c|}{4.88}                                                 & \multicolumn{1}{c|}{70.87\%} & 92.29\% & \multicolumn{1}{c|}{29.00\%}                                             & \multicolumn{1}{c|}{12.08\%}                                               & \multicolumn{1}{c|}{19.13}                                                & \multicolumn{1}{c|}{54.95\%} & 81.21\% \\ \hline
DSM (all) + AL + FI & 0.5                                                    & \multicolumn{1}{c|}{86.46\%}                                             & \multicolumn{1}{c|}{80.43\%}                                               & \multicolumn{1}{c|}{4.94}                                                 & \multicolumn{1}{c|}{70.48\%} & 92.23\% & \multicolumn{1}{c|}{28.14\%}                                             & \multicolumn{1}{c|}{12.00\%}                                               & \multicolumn{1}{c|}{18.82}                                                & \multicolumn{1}{c|}{55.58\%} & 81.75\% \\ \hline
DSM (all) + CS      & -                                                      & \multicolumn{1}{c|}{86.47\%}                                             & \multicolumn{1}{c|}{80.36\%}                                               & \multicolumn{1}{c|}{4.79}                                                 & \multicolumn{1}{c|}{71.23\%} & 92.39\% & \multicolumn{1}{c|}{27.27\%}                                             & \multicolumn{1}{c|}{11.44\%}                                               & \multicolumn{1}{c|}{19.17}                                                & \multicolumn{1}{c|}{54.06\%} & 80.77\% \\ \hline
DSM (all) + AL + CS & 0.1                                                    & \multicolumn{1}{c|}{86.18\%}                                             & \multicolumn{1}{c|}{80.19\%}                                               & \multicolumn{1}{c|}{4.81}                                                 & \multicolumn{1}{c|}{71.16\%} & 92.32\% & \multicolumn{1}{c|}{26.86\%}                                             & \multicolumn{1}{c|}{11.49\%}                                               & \multicolumn{1}{c|}{20.05}                                                & \multicolumn{1}{c|}{51.14\%} & 79.21\% \\ \hline
DSM (all) + AL + CS & 0.3                                                    & \multicolumn{1}{c|}{86.29\%}                                             & \multicolumn{1}{c|}{80.23\%}                                               & \multicolumn{1}{c|}{4.75}                                                 & \multicolumn{1}{c|}{72.28\%} & 92.43\% & \multicolumn{1}{c|}{28.45\%}                                             & \multicolumn{1}{c|}{11.78\%}                                               & \multicolumn{1}{c|}{18.61}                                                & \multicolumn{1}{c|}{54.13\%} & 81.73\% \\ \hline
DSM (all) + AL + CS & 0.5                                                    & \multicolumn{1}{c|}{86.38\%}                                             & \multicolumn{1}{c|}{80.17\%}                                               & \multicolumn{1}{c|}{4.81}                                                 & \multicolumn{1}{c|}{70.81\%} & 92.41\% & \multicolumn{1}{c|}{26.73\%}                                             & \multicolumn{1}{c|}{10.84\%}                                               & \multicolumn{1}{c|}{19.82}                                                & \multicolumn{1}{c|}{55.60\%} & 81.07\% \\ \hline
                    &                                                        & \multicolumn{5}{c|}{CVUSA -\textgreater CVUSA}                                                                                                                                                                                                                             & \multicolumn{5}{c|}{CVUSA -\textgreater CVACT}                                                                                                                                                                                                                             \\ \hline
DSM~\cite{ShiEtAl2020}*                 & -                                                      & \multicolumn{1}{c|}{93.29\%}                                             & \multicolumn{1}{c|}{80.54\%}                                               & \multicolumn{1}{c|}{5.29}                                                 & \multicolumn{1}{c|}{47.49\%} & 93.25\% & \multicolumn{1}{c|}{38.11\%}                                             & \multicolumn{1}{c|}{18.43\%}                                               & \multicolumn{1}{c|}{15.48}                                                & \multicolumn{1}{c|}{32.45\%} & 72.18\% \\ \hline
DSM (all)           & -                                                      & \multicolumn{1}{c|}{94.99\%}                                             & \multicolumn{1}{c|}{84.30\%}                                               & \multicolumn{1}{c|}{4.53}                                                 & \multicolumn{1}{c|}{50.76\%} & 95.01\% & \multicolumn{1}{c|}{41.40\%}                                             & \multicolumn{1}{c|}{20.03\%}                                               & \multicolumn{1}{c|}{13.90}                                                & \multicolumn{1}{c|}{35.08\%} & 75.82\% \\ \hline
DSM (all) + AL      & 0.1                                                    & \multicolumn{1}{c|}{94.75\%}                                             & \multicolumn{1}{c|}{84.47\%}                                               & \multicolumn{1}{c|}{4.75}                                                 & \multicolumn{1}{c|}{48.39\%} & 94.15\% & \multicolumn{1}{c|}{39.03\%}                                             & \multicolumn{1}{c|}{18.32\%}                                               & \multicolumn{1}{c|}{13.82}                                                & \multicolumn{1}{c|}{32.98\%} & 74.38\% \\ \hline
DSM (all) + AL      & 0.3                                                    & \multicolumn{1}{c|}{94.90\%}                                             & \multicolumn{1}{c|}{84.76\%}                                               & \multicolumn{1}{c|}{4.76}                                                 & \multicolumn{1}{c|}{47.20\%} & 93.67\% & \multicolumn{1}{c|}{41.15\%}                                             & \multicolumn{1}{c|}{20.32\%}                                               & \multicolumn{1}{c|}{13.99}                                                & \multicolumn{1}{c|}{33.17\%} & 74.01\% \\ \hline
DSM (all) + AL      & 0.5                                                    & \multicolumn{1}{c|}{94.82\%}                                             & \multicolumn{1}{c|}{84.87\%}                                               & \multicolumn{1}{c|}{5.03}                                                 & \multicolumn{1}{c|}{46.14\%} & 93.05\% & \multicolumn{1}{c|}{40.61\%}                                             & \multicolumn{1}{c|}{19.50\%}                                               & \multicolumn{1}{c|}{13.50}                                                & \multicolumn{1}{c|}{33.70\%} & 75.10\% \\ \hline
DSM (all) +FI       & -                                                      & \multicolumn{1}{c|}{95.30\%}                                             & \multicolumn{1}{c|}{84.91\%}                                               & \multicolumn{1}{c|}{3.77}                                                 & \multicolumn{1}{c|}{81.04\%} & 96.71\% & \multicolumn{1}{c|}{41.58\%}                                             & \multicolumn{1}{c|}{19.44\%}                                               & \multicolumn{1}{c|}{13.34}                                                & \multicolumn{1}{c|}{52.20\%} & 79.56\% \\ \hline
DSM (all) + AL + FI & 0.1                                                    & \multicolumn{1}{c|}{95.32\%}                                             & \multicolumn{1}{c|}{85.21\%}                                               & \multicolumn{1}{c|}{3.85}                                                 & \multicolumn{1}{c|}{81.21\%} & 96.63\% & \multicolumn{1}{c|}{42.97\%}                                             & \multicolumn{1}{c|}{20.67\%}                                               & \multicolumn{1}{c|}{12.93}                                                & \multicolumn{1}{c|}{52.62\%} & 79.47\% \\ \hline
DSM (all) + AL + FI & 0.3                                                    & \multicolumn{1}{c|}{95.29\%}                                             & \multicolumn{1}{c|}{84.98\%}                                               & \multicolumn{1}{c|}{3.78}                                                 & \multicolumn{1}{c|}{82.13\%} & 96.77\% & \multicolumn{1}{c|}{41.93\%}                                             & \multicolumn{1}{c|}{20.07\%}                                               & \multicolumn{1}{c|}{13.15}                                                & \multicolumn{1}{c|}{52.94\%} & 79.77\% \\ \hline
DSM (all) + AL + FI & 0.5                                                    & \multicolumn{1}{c|}{95.35\%}                                             & \multicolumn{1}{c|}{85.28\%}                                               & \multicolumn{1}{c|}{3.84}                                                 & \multicolumn{1}{c|}{82.15\%} & 96.76\% & \multicolumn{1}{c|}{42.02\%}                                             & \multicolumn{1}{c|}{20.01\%}                                               & \multicolumn{1}{c|}{13.30}                                                & \multicolumn{1}{c|}{52.17\%} & 79.39\% \\ \hline
DSM (all) + CS      & -                                                      & \multicolumn{1}{c|}{95.32\%}                                             & \multicolumn{1}{c|}{84.84\%}                                               & \multicolumn{1}{c|}{3.73}                                                 & \multicolumn{1}{c|}{81.80\%} & 96.79\% & \multicolumn{1}{c|}{42.71\%}                                             & \multicolumn{1}{c|}{20.59\%}                                               & \multicolumn{1}{c|}{12.77}                                                & \multicolumn{1}{c|}{53.83\%} & 80.14\% \\ \hline
DSM (all) + AL + CS & 0.1                                                    & \multicolumn{1}{c|}{95.19\%}                                             & \multicolumn{1}{c|}{84.93\%}                                               & \multicolumn{1}{c|}{3.74}                                                 & \multicolumn{1}{c|}{82.20\%} & 96.78\% & \multicolumn{1}{c|}{42.78\%}                                             & \multicolumn{1}{c|}{20.34\%}                                               & \multicolumn{1}{c|}{12.92}                                                & \multicolumn{1}{c|}{52.36\%} & 79.88\% \\ \hline
DSM (all) + AL + CS & 0.3                                                    & \multicolumn{1}{c|}{95.27\%}                                             & \multicolumn{1}{c|}{85.19\%}                                               & \multicolumn{1}{c|}{3.77}                                                 & \multicolumn{1}{c|}{82.42\%} & 96.75\% & \multicolumn{1}{c|}{42.37\%}                                             & \multicolumn{1}{c|}{20.00\%}                                               & \multicolumn{1}{c|}{13.02}                                                & \multicolumn{1}{c|}{54.30\%} & 80.21\% \\ \hline
DSM (all) + AL + CS & 0.5                                                    & \multicolumn{1}{c|}{95.29\%}                                             & \multicolumn{1}{c|}{85.24\%}                                               & \multicolumn{1}{c|}{3.84}                                                 & \multicolumn{1}{c|}{81.84\%} & 96.74\% & \multicolumn{1}{c|}{42.40\%}                                             & \multicolumn{1}{c|}{20.24\%}                                               & \multicolumn{1}{c|}{12.85}                                                & \multicolumn{1}{c|}{52.30\%} & 80.05\% \\ \hline
\end{tabular}
\end{table*}

\subsection{Best and worst example visualization} \label{session: visual_samples}
\begin{figure}[!h]
\vspace{-1em}
  \includegraphics[width=0.92 \textwidth, trim = 1cm 9.7cm 28.8cm 0.5cm, clip]{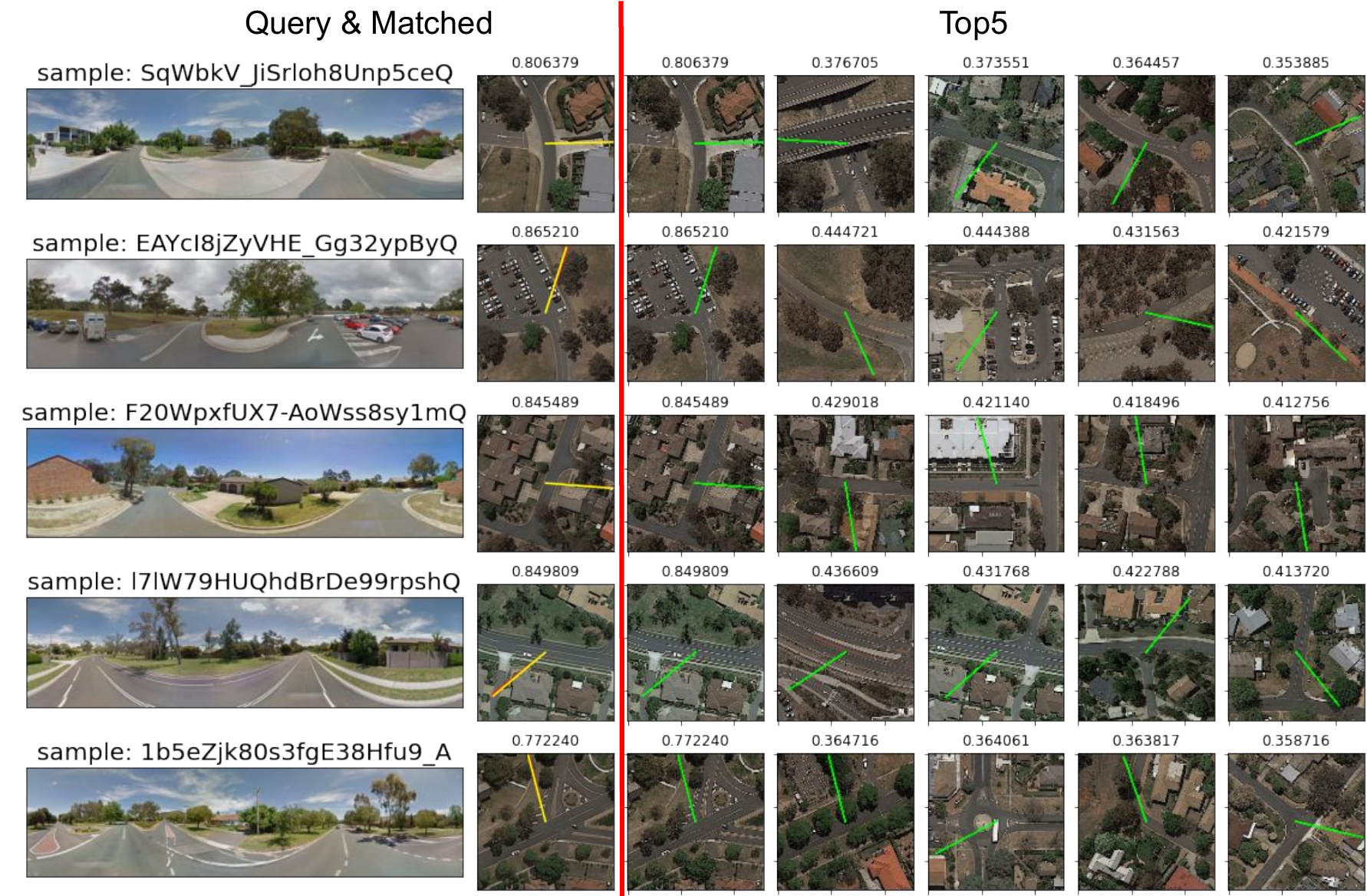}
  \vspace{-1em}
  \caption{Top best samples from CVACT~\cite{LiuLi2019} FI model with highest similarity margins.}
  \label{fig:cvact fi best}
\end{figure}

\begin{figure}[!h]
\vspace{-1em}
  \includegraphics[width=0.92 \textwidth, trim = 1cm 9.7cm 28.8cm 0.5cm, clip]{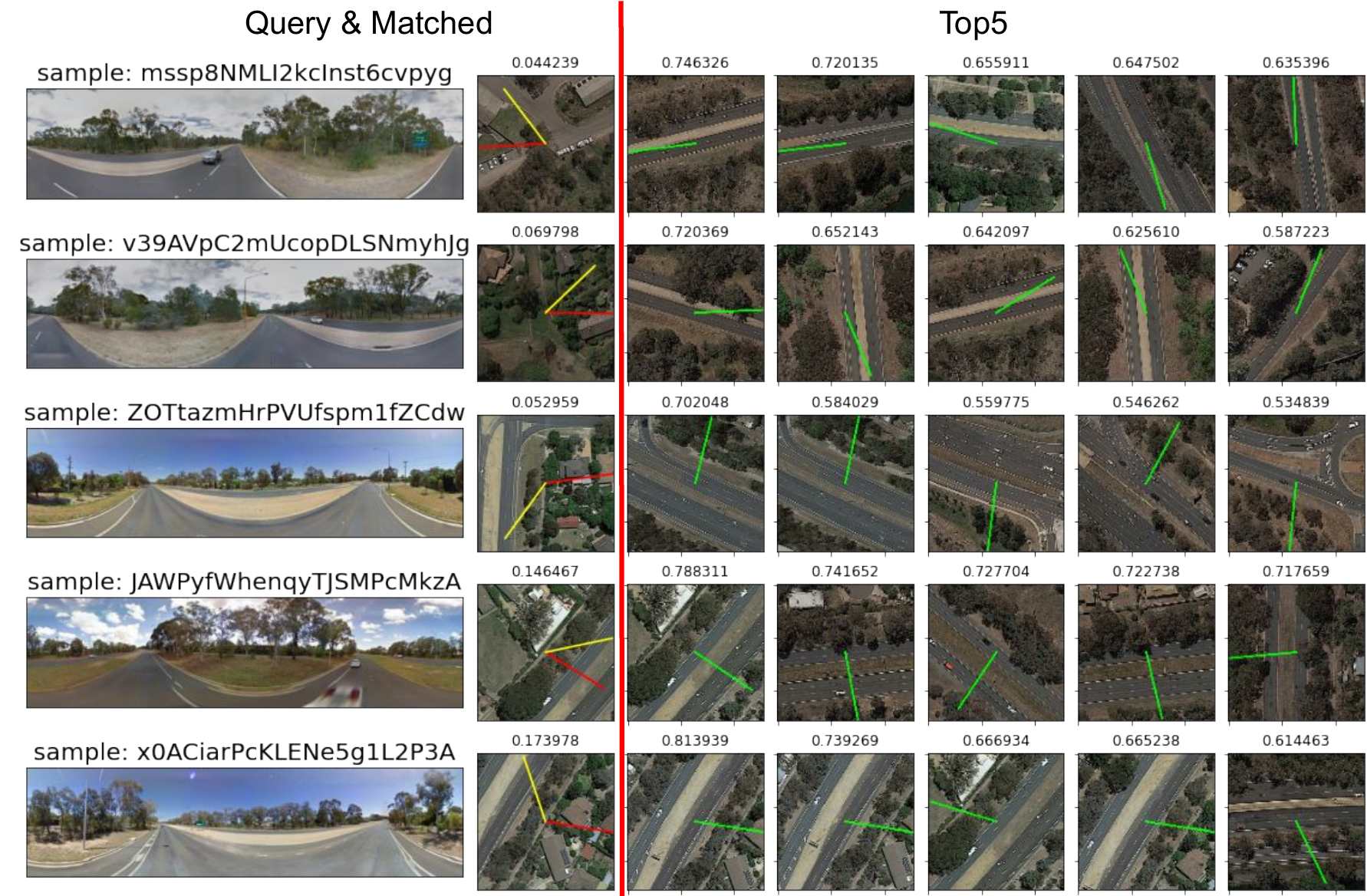}
  \vspace{-1em}
  \caption{Top worst samples from CVACT~\cite{LiuLi2019} FI model with highest similarity margins.}
  \label{fig:cvact fi worst}
  \vspace{-1em}
\end{figure}

 \multiref{fig:cvact fi best}{fig:cvusa dsm worst} show the visualization of the best matched cases (top 5 correctly matched queries with the highest margin between the similarity scores of the top~1 and the top~2 prediction) and the worst mismatched cases (top 5 wrongly matched queries with the highest margin between the similarity scores of top~1 prediction and the correct match). Each row shows the street-view query, matched satellite image (yellow line indicates the extracted orientation, the red line indicates the ground truth), Top 5 prediction results (green line indicates the extracted orientation) with similarity scores. Note that in \autoref{fig:cvact fi worst}, all five samples have GPS drift, the actual camera locations are not at the center of the ground truth images. Besides the third row, the other four worst samples actually find better-matched images (the top1 images are closer to the correct location) than the ground truth images and the orientation estimations are correct. 

\begin{figure}[!h]
\vspace{-1em}
  \includegraphics[width=0.92 \textwidth, trim = 1cm 9.7cm 28.8cm 0.5cm, clip]{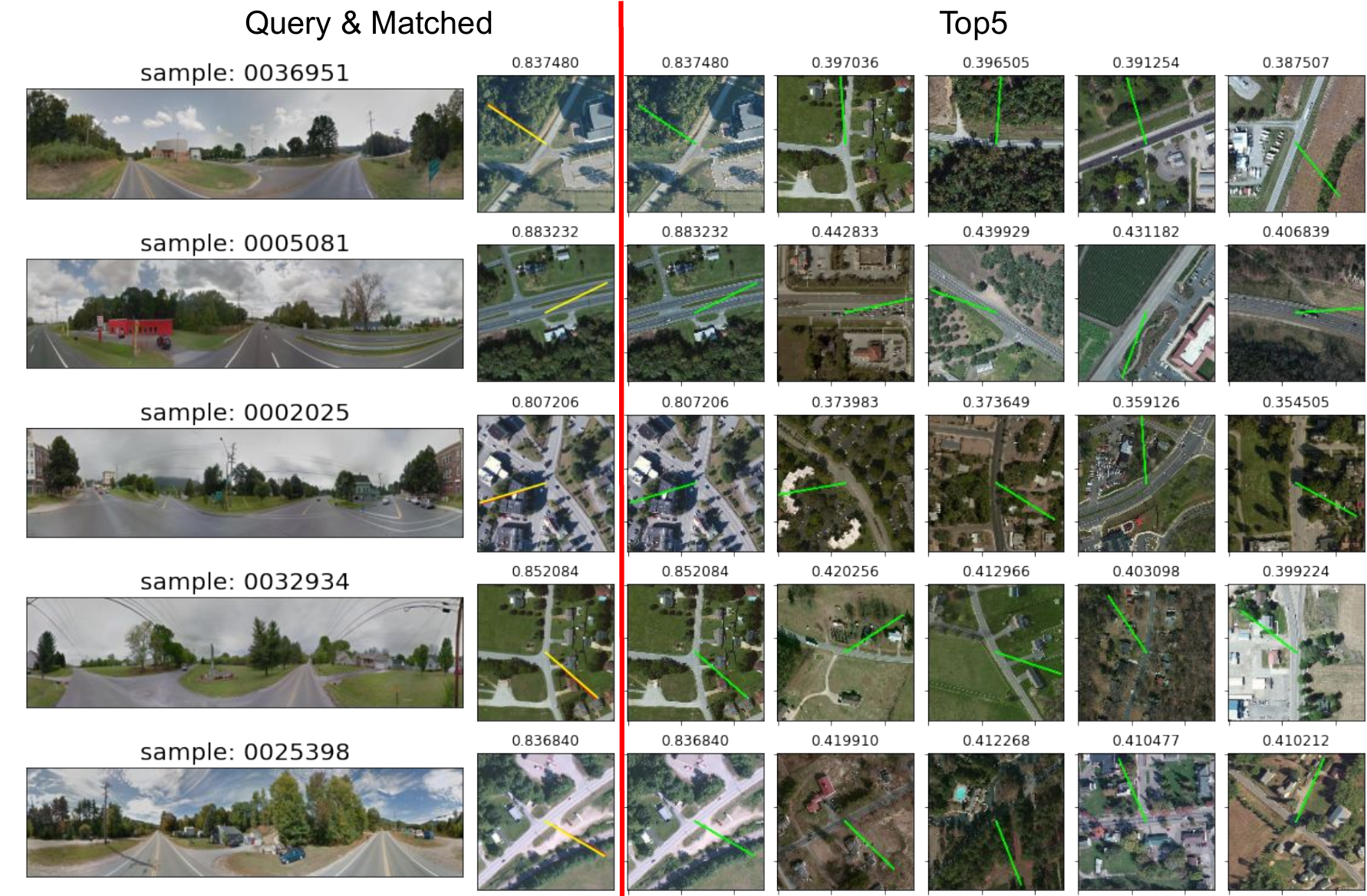}
  \vspace{-1em}
  \caption{Top best samples from CVUSA~\cite{WorkmanEtAl2015} CS model with highest similarity margins.}
  \label{fig:cvusa cs best}
\end{figure}

\begin{figure}[!h]
\vspace{-1em}
  \includegraphics[width=0.92 \textwidth, trim = 1cm 9.7cm 28.8cm 0.5cm, clip]{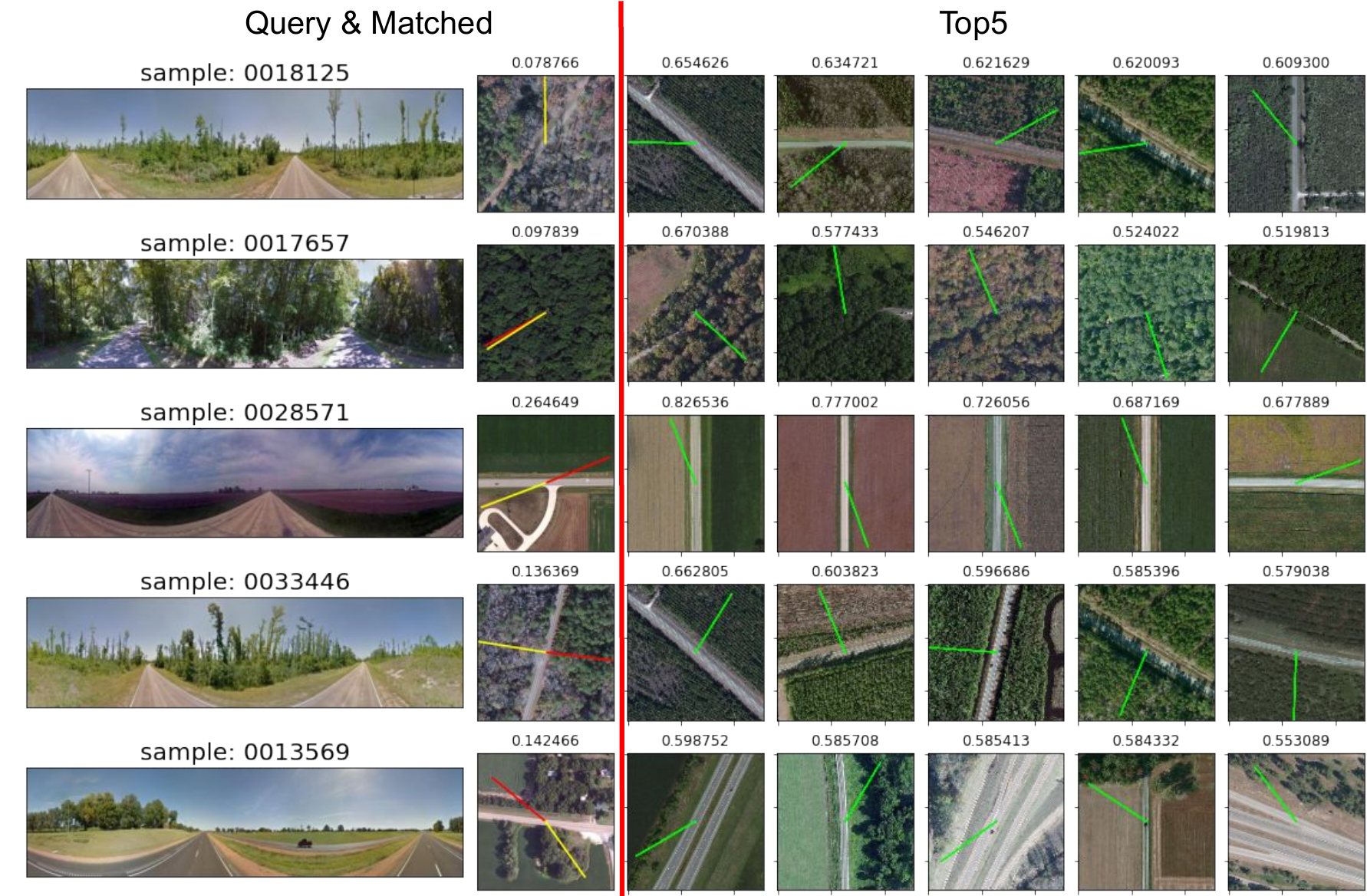}
  \vspace{-1em}
  \caption{Top worst samples from CVUSA~\cite{WorkmanEtAl2015} CS model with highest similarity margins.}
  \label{fig:cvusa cs worst}
  \vspace{-1em}

\end{figure}

\begin{figure}[!h]
\vspace{-1em}
  \includegraphics[width=0.92 \textwidth, trim = 1cm 9.7cm 28.8cm 0.5cm, clip]{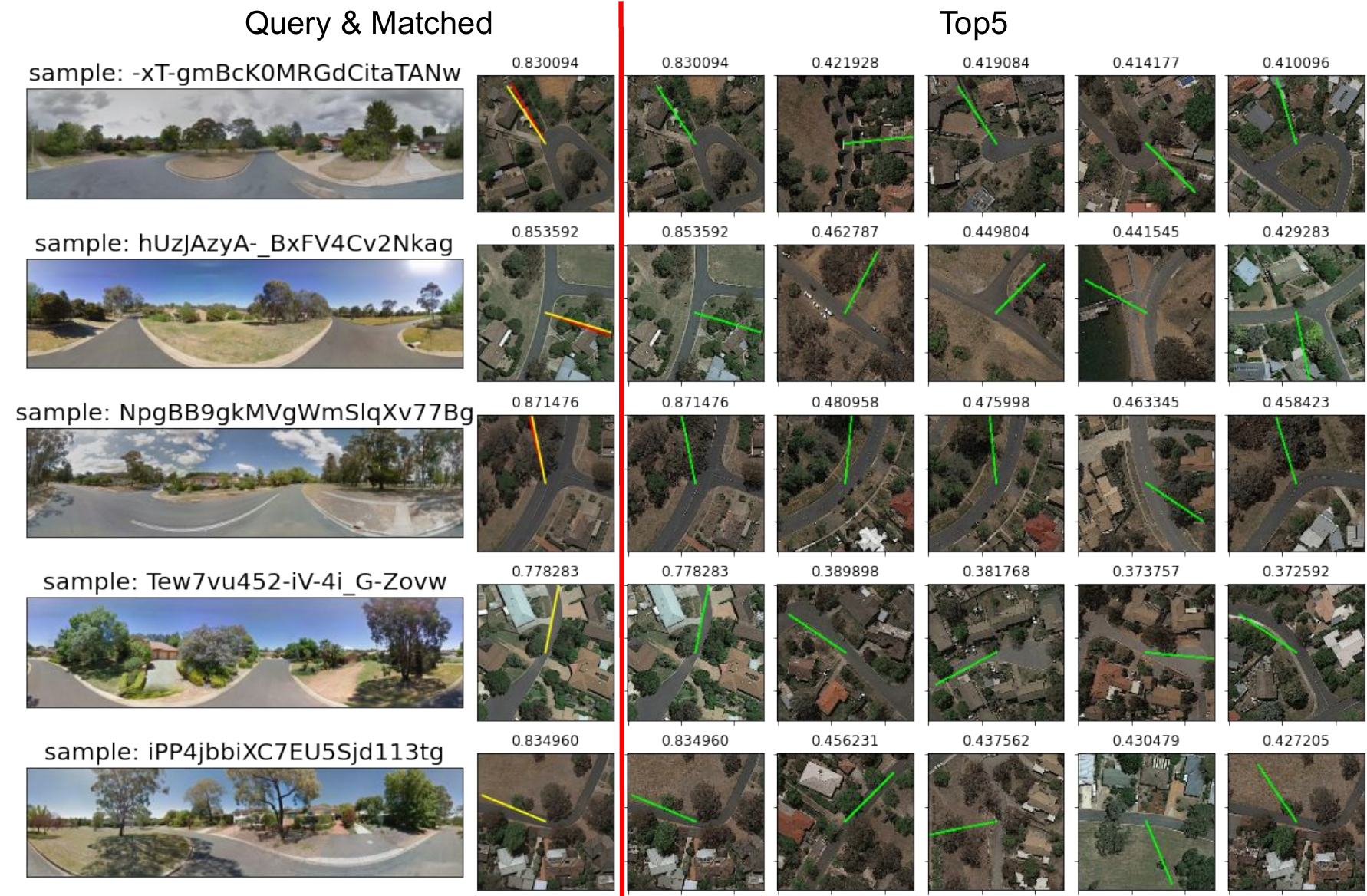}
  \vspace{-1em}
  \caption{Top best samples from CVACT~\cite{LiuLi2019} DSM~\cite{ShiEtAl2020}* model with highest similarity margins.}
  \label{fig:cvact dsm best}
  
\end{figure}

\begin{figure}[!h]
\vspace{-1em}
  \includegraphics[width=0.92 \textwidth, trim = 1cm 9.7cm 28.8cm 0.5cm, clip]{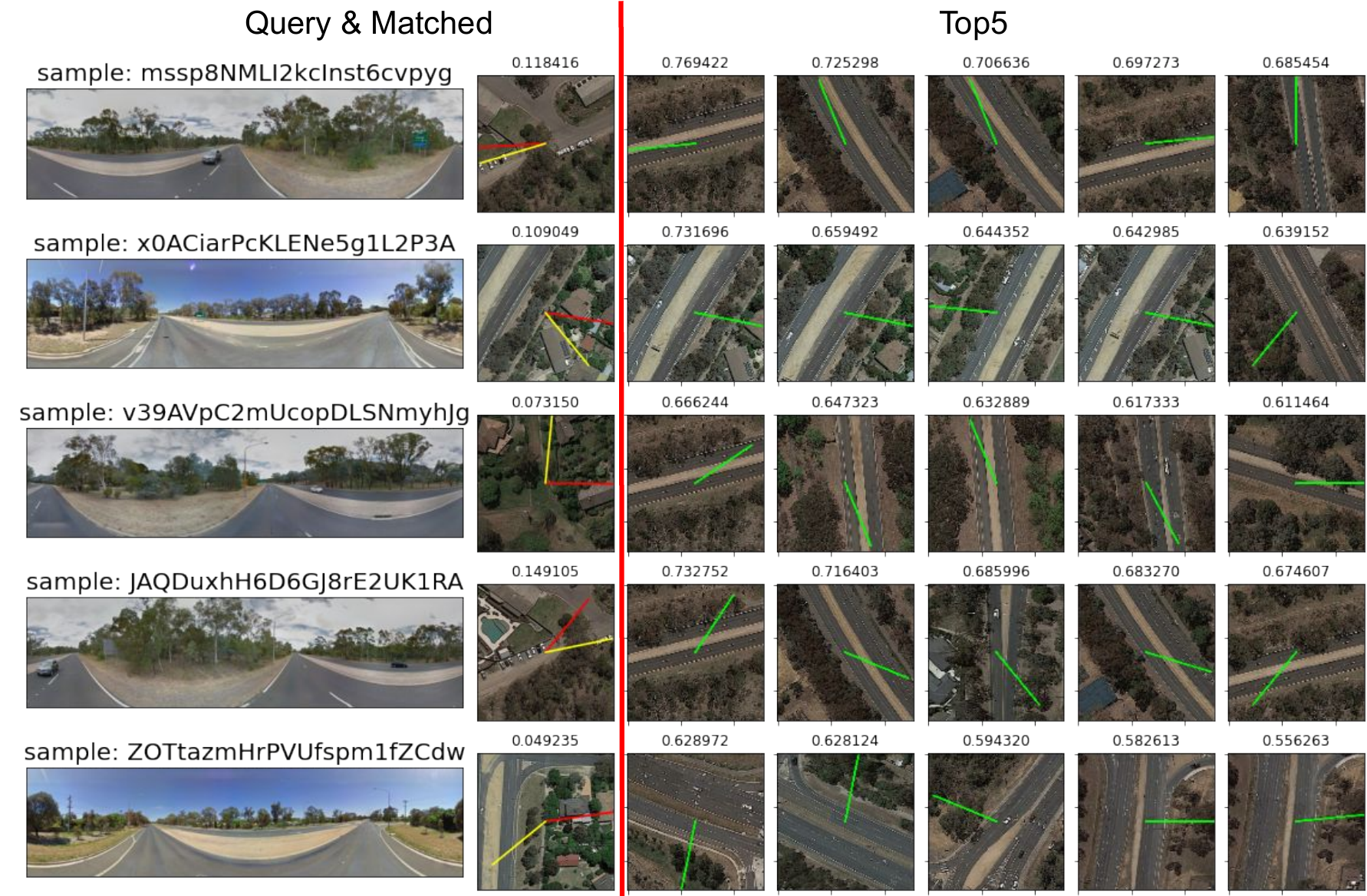}
  \vspace{-1em}
  \caption{Top worst samples from CVACT~\cite{LiuLi2019} DSM~\cite{ShiEtAl2020}* model with highest similarity margins.}
  \label{fig:cvact dsm worst}
  \vspace{-1em}
\end{figure}

\begin{figure}[!h]
\vspace{-1em}
  \includegraphics[width=0.92 \textwidth, trim = 1cm 9.7cm 28.8cm 0.5cm, clip]{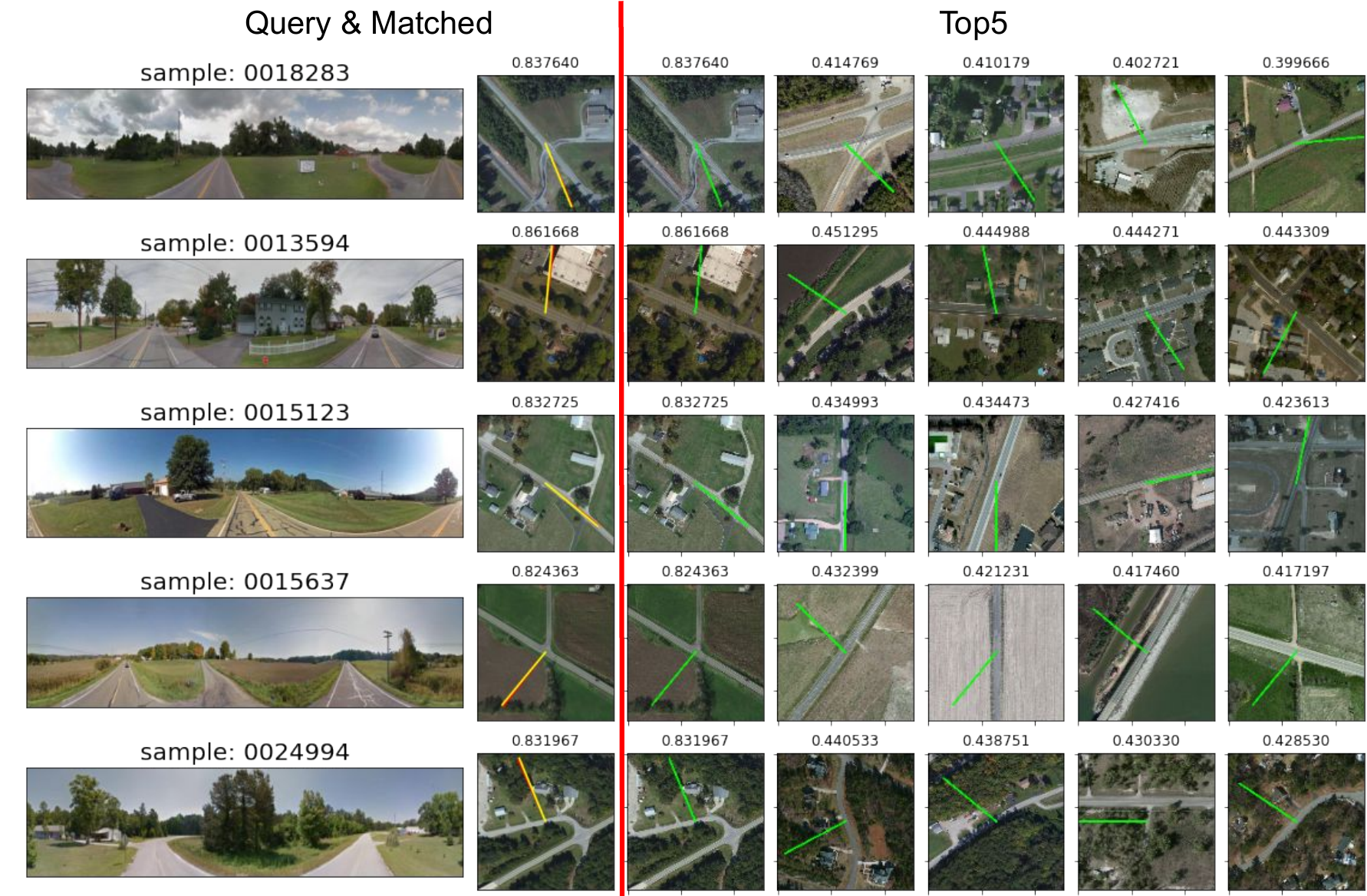}
  \vspace{-1em}
  \caption{Top best samples from CVUSA~\cite{WorkmanEtAl2015} DSM~\cite{ShiEtAl2020}* model with highest similarity margins.}
  \label{fig:cvusa dsm best}
  
\end{figure}

\begin{figure}[!h]
\vspace{-1em}
  \includegraphics[width=0.92 \textwidth, trim = 1cm 9.7cm 28.8cm 0.5cm, clip]{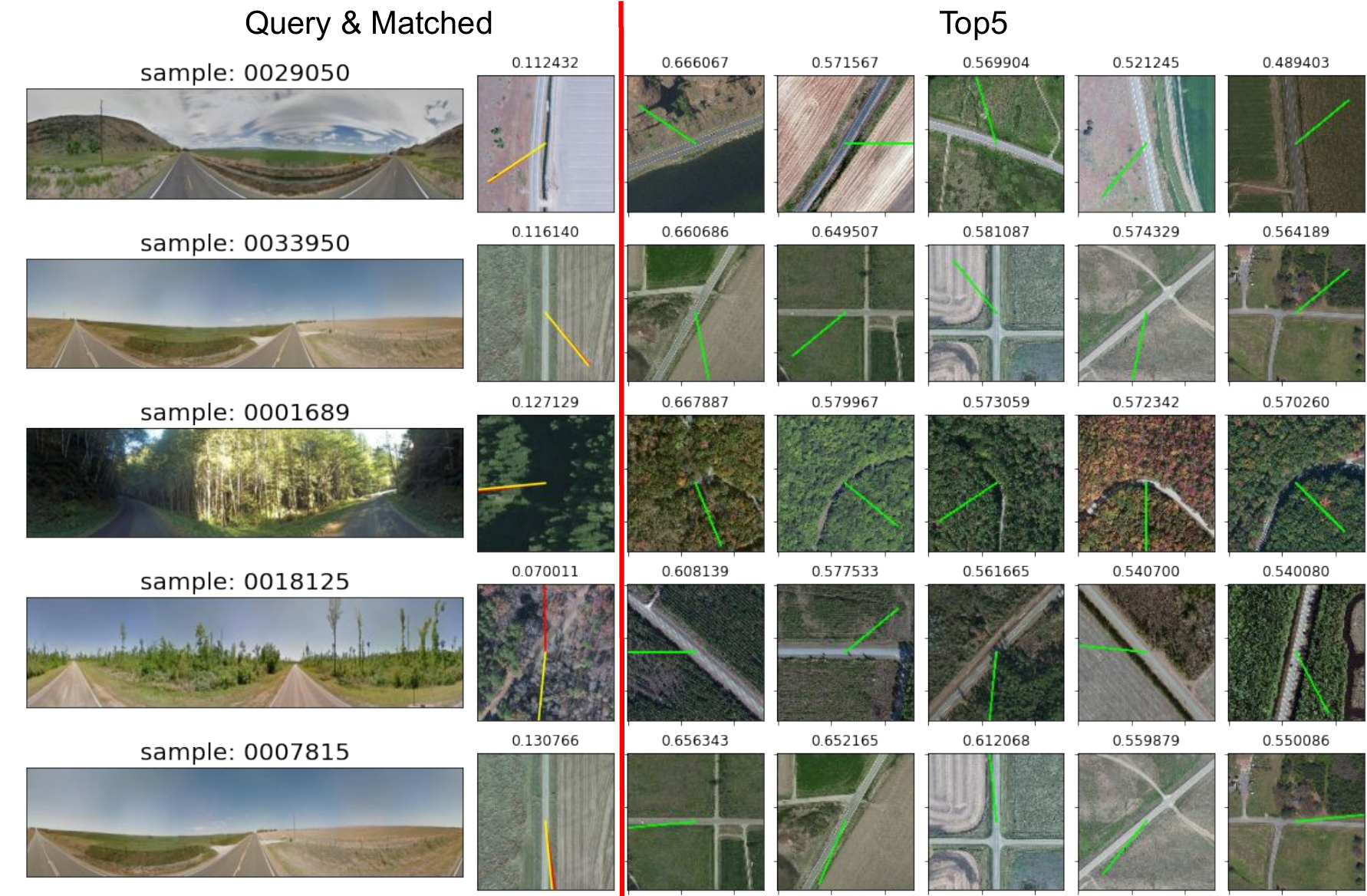}
  \vspace{-1em}
  \caption{Top worst samples from CVUSA~\cite{WorkmanEtAl2015} DSM~\cite{ShiEtAl2020}* model with highest similarity margins.}
  \label{fig:cvusa dsm worst}
  \vspace{-1em}
\end{figure}

\end{document}